\documentclass{article}

\usepackage{PRIMEarxiv}

\usepackage[utf8]{inputenc} 
\usepackage[T1]{fontenc}    
\usepackage{hyperref}       
\usepackage{url}            
\usepackage{booktabs}       
\usepackage{amsfonts}       
\usepackage{nicefrac}       
\usepackage{microtype}      
\usepackage{lipsum}
\usepackage{fancyhdr}       
\usepackage{graphicx}       
\graphicspath{{media/}}     
\usepackage{amsmath}
\usepackage{float}
\usepackage[caption=false,font=normalsize,labelfont=sf,textfont=sf]{subfig} 
\usepackage{textcomp}
\usepackage{stfloats}
\usepackage{url}
\usepackage{verbatim}
\usepackage{graphicx}
\usepackage{cite}
\usepackage{booktabs} 
\usepackage{xcolor}
\usepackage{amssymb}  
\usepackage{amsmath}
\usepackage{float}
\title{A two-stream network with global-local feature fusion for bone age assessment}

\author{
  Qiong Lou \\
  School of Science \\
  Zhejiang University of Science and Technology \\        
  Hangzhou,China \\ 
  \texttt{louqiong@zust.edu.cn} \\
   \And
  Han Yang \\
  School of Science \\
  Zhejiang University of Science and Technology \\        
  Hangzhou,China \\ 
  \texttt{yhan27@zust.edu.cn} \\
  \And
  Fang Lu$^{*}$ \\
  School of Science \\  
  Zhejiang University of Science and Technology \\       
  Hangzhou,China \\       
  \texttt{lufang@zust.edu.cn} \\
}

\begin{document}
\maketitle

\begin{abstract}
Bone Age Assessment (BAA) is a widely used clinical technique that can accurately reflect an individual's growth and development level, as well as maturity. In recent years, although deep learning has advanced the field of bone age assessment, existing methods face challenges in efficiently balancing global features and local skeletal details. This study aims to develop an automated bone age assessment system based on a two-stream deep learning architecture to achieve higher accuracy in bone age assessment. We propose the BoNet+ model incorporating global and local feature extraction channels. A Transformer module is introduced into the global feature extraction channel to enhance the ability in extracting global features through multi-head self-attention mechanism. A RFAConv module is incorporated into the local feature extraction channel to generate adaptive attention maps within multiscale receptive fields, enhancing local feature extraction capabilities. Global and local features are concatenated along the channel dimension and optimized by an Inception-V3 network. The proposed method has been validated on the Radiological Society of North America (RSNA) and Radiological Hand Pose Estimation (RHPE) test datasets, achieving mean absolute errors (MAEs) of 3.81 and 5.65 months, respectively. These results are comparable to the state-of-the-art. The BoNet+ model reduces the clinical workload and achieves automatic, high-precision, and more objective bone age assessment.
\end{abstract}


\section{Introduction}
Bone Age Assessment is of great significance in determining the skeletal development age of children. It helps with the prediction of growth potential, the diagnosis of precocious puberty, and the identification of endocrine disorders \cite{giordano2010automatic}. Traditionally, radiologists assess bone age by observing the maturity of the phalanges, carpal, radius, and ulna based on left hand X-ray images and comparing them with established standards such as the Greulich–Pyle (GP) atlas \cite{ref1} and the Tanner–Whitehouse (TW) scores  \cite{ref2,ref3,ref4}. The GP method is a template matching method, in which X-ray images are compared with annotated image sets and the matching results are used as the evaluation results. The GP method has been widely used internationally due to its simplicity, clarity, and ease of application. The TW method defines regions of interest (ROIs), the radiologist scores each ROI and then sums them up to determine the maturity of hand bone. However, both of these clinical methods rely heavily on the experience of radiologists, leading to marked inter-observer and intra-observer variability. Additionally, the ROIs annotation and bone age assessment are time-consuming and laborious. In recent years, with the remarkable progress in computational power and the development of deep learning techniques, many researchers have focused on developing rapid, accurate, and more objective BAA methods. Deep learning techniques have achieved remarkable success in various tasks within computer vision, medical image analysis, and biomedical fields. In the domain of BAA, the research methods mainly include traditional learning methods and deep learning methods.

Traditional learning methods require manual feature extraction. Due to the long time consumption, large errors, poor stability, and the inability to conduct fully automated evaluation, these methods are limited in clinical applications. In 2017, Spampinato et al. \cite{ref5} were the first to apply deep learning to the BAA field. They developed a convolutional neural network (CNN) integrated with deformable layers, which achieved state-of-the-art (SOTA) result at that time and marked a critical transition from shallow learning paradigm to deep learning paradigm in skeletal maturity evaluation. Subsequent deep learning methods can be classified into the black-box-based method and the ROI-based method. The black-box-based method typically involves a single end to end network that inputs the whole hand radiographs into a CNN to obtain predicted bone age \cite{ref5,ref6,ref7}. Although this method has a simple structure and is easy to deploy, it suffers from inherent limitations in model interpretability. The ROI-based method typically consists of two steps. Firstly, ROIs are detected and extracted by leveraging object detection networks like Faster R-CNN \cite{ref8} and YOLO \cite{ref9}. Subsequently, these ROIs are input into a BAA network. However, this method also has some limitations. Firstly, extracting ROIs requires precise manual annotations, which are time-consuming and laborious. Secondly, training the object detection network demands a long time, and minor errors in the detection process will be amplified in the final bone age prediction. Thirdly, focusing solely on local ROIs may lead to the loss of global information. To overcome these limitations, Escobar et al. \cite{ref10} proposed a BoNet network that utilizes the whole hand radiographs and attention maps to perform BAA. Attention maps generated from keypoint annotations guide the model to focus on important skeletal regions. Although this method reduces the need for ROI segmentation, it still suffers from insufficient coverage of keypoint regions. To address this issue, our method introduces a Receptive-Field Attention Convolutional (RFAConv) module \cite{ref11} in the local feature extraction channel. This module effectively compensates for the insufficient coverage of keypoint regions in attention maps, thereby enhancing local feature extraction capability. Additionally, in the global feature extraction channel, our method introduces a Transformer module \cite{ref12} that enables the model to fully focus on global information in the whole hand radiographs, thus capturing the overall development of the skeletal more accurately. Based on these improvements, we propose a method entitled BoNet+ that not only retains the advantages of the BoNet network \cite{ref10} but also further enhances the accuracy and robustness of BAA.

The main contributions of the proposed method are summarized as follows:

1.Global feature extraction based on Transformer. A Transformer module is introduced in the global feature extraction channel. This module enables the model to better understand the overall skeletal development of the whole hand radiographs, and also enhances the model's grasp of overall skeletal maturity. The feature interested here is similar to that of the GP method.

2. Local feature extraction based on RFAConv. A RFAConv module is introduced in the local feature extraction channel. This module effectively compensates for the insufficient coverage of keypoint regions in attention maps. It enables the model to more comprehensively focus on the morphological features of the distal phalanges, middle phalanges, proximal phalanges, and metacarpal bones. Additionally, it enhances the recognition of new bone growth in the carpal region. The feature interested here is similar to that of the TW method.

3. Network performance evaluation. Experiments conducted on two datasets demonstrate that the proposed method not only performs well in bone age assessment but also offers valuable auxiliary information to support the clinical process. 

\section{Related Work}\label{sec:related}
With remarkable progress in computational capabilities and the continuous evolution of deep learning techniques, the black-box-based BAA methods have emerged as mainstream methods, which have driven the transition of bone age assessment from manual to automated. These methods are able to assess bone age in a single step. Chen [11] adopted the GP method framework and proposed a BAA classification model based on transfer learning and VGGNet \cite{ref13}, obtaining MAE of 13.2 months on the DHA dataset \cite{gertych2007bone}. Spampinato et al. \cite{ref5} tested multiple CNNs, including OverFeat \cite{ref14}, GoogLeNet \cite{ref15}, and VGGNet, and developed an automated system that reached SOTA MAE of 9.48 months on the DHA dataset. This achievement represented a significant shift from the shallow learning paradigm to the deep learning paradigm in automated BAA research. Lee and Kim \cite{ref7} designed a network composed of convolutional layers, pooling layers, and fully connected layers, and introduced an image preprocessing procedure to reduce the impact of irrelevant regions, achieving MAE of 18.9 months on the RSNA test dataset. Larson et al. \cite{ref16} utilized ResNet \cite{ref17} to extract features from the whole hand radiographs, achieving MAE of 6.24 months on the RSNA test dataset. Tang et al. \cite {ref18} used ResNet50 \cite {ref17} as the backbone network and added specific networks with convolutional and fully connected layers, which were designed to extract high level features specific for males and females. This method obtained MAE of 5.53 months on the RSNA validation dataset. Although the black-box-based methods have simplified the BAA process, inputting the whole hand radiographs into the CNN network for BAA results in a lack of interpretability. Consequently, researchers have begun to focus on key skeletal regions of the hand.

The ROI-based methods generally consist of two steps. The first step is to localize the ROIs in the hand, which can be categorized into two main methods: (i) detecting and segmenting specific ROIs using detection networks, and (ii) localizing ROIs through attention learning. The second step is to input the ROIs into a BAA network, which can be either a CNN or a Transformer network. 

In the following, we first introduce methods that use detection networks to obtain specific ROIs and then employ CNNs for predicting bone age. Bui et al. \cite{ref20} used Faster R-CNN \cite{ref8} to detect seven ROIs, and then used Inception-v4 \cite{ref21} for bone age classification, achieving MAE of 7.08 months on the DHA dataset. Koitka et al. \cite{ref22} proposed an ossification region detection network to detect and extract ROIs, and then used a region-specific regression network for bone age prediction, obtaining MAE of 4.56 months on the RSNA test dataset. Ji et al. \cite{ref23} developed a module to measure the importance of different ROIs and selected the highest ranked ROIs to aid the algorithm. Then, they used ResNet34 \cite{ref17} for the prediction of bone age. This method obtained MAE of 4.49 months on the RSNA test dataset. Li et al. \cite{ref24} introduced a regional aggregation graph convolutional network (RAGCN) that combines the strengths of CNN and graph convolutional networks (GCN). First, GCN was employed to infer the key bone regions. Subsequently, CNN channels were designated for each ROI to extract image features. The RAGCN then fused the features of these ROIs and ultimately integrated its output with that of the CNN to generate the final bone age prediction. This method achieved MAE of 4.09 months on the RSNA test dataset and 6.78 months on the RHPE test dataset. Liu et al. \cite{ref25} proposed a model with a self supervised mechanism that can be trained end to end. Under weak supervision, the network discovered the most discriminative ROIs through inter-region and intra-region consistency. Then, a region extraction and age identification network was used to predict bone age. This method achieved MAE of 3.99 months on the RSNA test dataset. The significance of the ROI-based methods lies in their direct alignment with clinical diagnostic logic. By focusing on anatomically meaningful regions, these approaches bridge the gap between automated algorithms and radiological practice, enabling interpretable feature extraction that mirrors how clinicians assess skeletal maturity. However, methods using detection networks to segment specific ROIs have limitations. The hand bones have a complex structure, during childhood skeletal development, there are significant individual differences. This increases the difficulty for detection networks to accurately segment or localize certain complex regions. As a result, it may affect the accuracy of bone age prediction. Moreover, all the above mentioned methods rely on multiple ROIs. Excluding certain key ROIs may also affect the accuracy of bone age prediction.

In contrast, the attention-learning-based methods can automatically acquire these ROIs without additional prior information. Ren et al. \cite {ref26} first used an attention module to process all images or generate coarse and fine attention maps. These maps were then input into an Inception-V3 \cite {ref27} network for bone age prediction. This method achieved MAE of 5.2 months on the RSNA test dataset. Chen et al. \cite{ref28} first used an attention guided approach to automatically localize discriminative regions without any additional annotations. They further employed joint age distribution learning and expectation regression, leveraging the ordinal relationships between individuals' ages, to enhance the robustness of age estimation. This method achieved MAE of 4.4 months on the RSNA test dataset. Wang et al. \cite{CFJLNet} proposed a coarse and fine feature joint learning network that employs dual branches to generate coarse and fine attention maps. The network integrates a dual feature fusion module to capture long-term dependencies in attention and an attention in attention module to guide fine attention generation via coarse attention. This approach achieved a MAE of 4.07 months on the RSNA test dataset. Escobar et al. \cite{ref10} proposed a bone age prediction method using hand bone keypoints as auxiliary information. They annotated 17 keypoints in the hand bones and created attention maps by generating Gaussian distributions around each keypoint, and then used the whole hand radiographs and attention maps for BAA. This method achieved MAE of 4.14 months on the RSNA test dataset and 7.6 months on the RHPE test dataset. However, the limited coverage of keypoints in their attention maps resulted in insufficient skeletal information. To solve this issue, we introduce a RFAConv module in the local feature extraction channel to compensate for the insufficient coverage of keypoints in attention maps. González et al. \cite{ref29}, building on the BoNet network [10], incorporated patients' chronological ages as prior information. They proposed a new task of predicting the deviation between bone age and chronological age instead of directly predicting bone age, arguing that this approach better aligns with clinical workflows. This method achieved MAE of 5.47 months on the RHPE test dataset. Liu et al. \cite{ref30} proposed a novel network called rich attention network (RA-Net), with a baseline network of ResNet50 for feature extraction from input images. RA-Net incorporated a lightweight rich attention module to generate a set of attention maps from the baseline network's feature maps. Each attention map was element wise multiplied with the original feature map to produce enhanced feature maps, ensuring the model focuses on key anatomical features. This method obtained MAE of 4.10 months on the RSNA test dataset. Wu et al. \cite{ref31} proposed a novel network named BoGFF-Net. Initially, it fused feature maps from different layers via a multi-scale self attention mechanism. Subsequently, an adaptive triplet loss function was introduced to distinguish sample pairs during regression tasks. Eventually, a multi layer perceptron (MLP) was employed to carry out bone age prediction. This method achieved MAE of 3.91 months on the RSNA test dataset and 7.07 months on the DHA dataset. Yang et al. \cite{ref32} used ResNet as the base architecture. They inserted multi-scale multi-receptive complementary attention modules after each convolutional block to generate texture features, and then used a graph attention module to generate contextual features, which were combined with encoded gender features. These combined features were input into MLP for the final bone age prediction. This method obtained MAE of 3.88 months on the RSNA test dataset. Compared to methods using detection networks to segment specific ROIs, the core significance of the attention-learning-based ROI localization methods lies in breaking through traditional annotation dependence. These methods enable neural networks to autonomously learn discriminative regions of skeletal development, avoiding the reliance of detection networks on precise manual segmentation and consuming less time.

Recently, the Transformer architecture \cite {ref33} has found extensive applications in the fields of computer vision and medical image analysis. DETR \cite {ref34} was the first model to apply the Transformer to the computer vision field, mainly for object detection tasks. Subsequently, Vision Transformer (ViT) \cite{ref12} gradually gained attention as an image classification model. It followed the traditional Transformer architecture by segmenting images into fixed size patches and models global information in images through self attention mechanisms. In the field of medical image analysis, the Transformer architecture has also been introduced into BAA tasks. Mao et al. \cite{ref35} proposed a two-stage convolutional transformer network based on the whole hand radiographs. They used the YOLOv5 network \cite{ref9} to detect and segment 18 ROIs, and then used the Swin Transformer \cite{ref36} network to extract feature information between ROIs for bone age prediction. This method achieved MAE of 4.586 months on the RSNA test dataset. Wu et al. \cite{ref37} developed an automatic bone age assessment network based on the TW3 method and ViT. First, a pre-trained YOLO network was used to remove the arm part, and a spatial configuration network was used to locate 37 keypoints of hand bones. The image was then segmented into 20 fixed size image patches, each of which was classified through ViT. This method was validated using a private dataset and achieved MAE of 5.64 months. Zhao et al. \cite{ref38} used the U-net network \cite{ref39} for ROI localization and segmentation, and used a pre-trained ResNet18 network \cite{ref17} to extract features from each ROI. These feature vectors were input into the Transformer encoder block to learn the similarity and correlation between ROIs, resulting in 13 output features of ROIs. The Inception ResNetV2 \cite{ref21} network was then used to extract global features from the entire image, the local and global features were concatenated and input into MLP to output the predicted bone age. This method obtained MAE of 4.56 months on the RSNA test dataset. Based on the remarkable global information capturing advantages of the Transformer in BAA tasks demonstrated above, this module enables the model to effectively integrate long-range spatial dependencies across hand bone structures, thereby enhancing the accuracy of skeletal maturity assessment. We introduce a Transformer module in the global feature extraction channel. This addition enhances the model's ability to extract global information, leading to a more comprehensive understanding of the overall structure and development of hand bones. 

In recent years, the clinical translation of bone age assessment systems has made remarkable progress, with deep learning-driven solutions entering practical medical application scenarios. A typical case is the uAI Discover BoneAge system \cite{United Imaging Intelligence}, which not only supports multi-protocol evaluation, but also integrates physiological parameters to enable comprehensive analysis of growth trajectory. The system has reduced the MAE to 4 months. Such applications indicate that deep learning methods are no longer confined to research benchmarks, they are effectively addressing clinical assessment efficiency, diagnostic consistency, and multi-parameter integration questions. The successful use of the uAI system demonstrates that the combination of algorithmic advancements and clinical experience can bring substantial improvements to pediatric medical services.

\section{Materials and Methods}\label{sec:method}

\begin{figure*}[htbp]
    \centering
    \includegraphics[width=0.8\textwidth, trim={220.225bp 0 220.225bp 0}, clip]{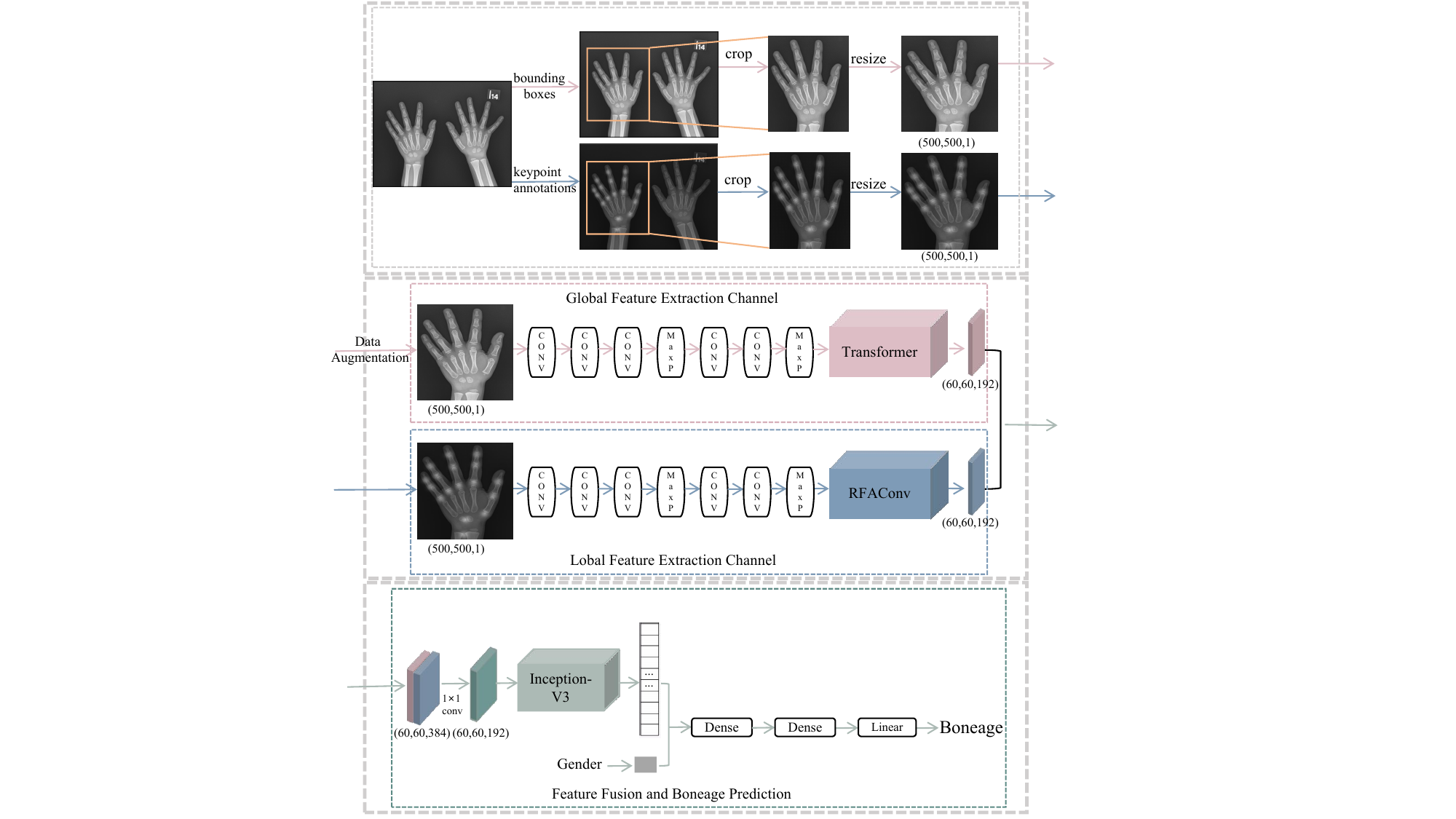}
    \caption{Overview of the pipeline used in BoNet+.}
    \label{fig:my_label}
\end{figure*}

In this article, we present an enhanced method named BoNet+. The architecture of this network is visually depicted in Fig.\ref{fig:my_label}. First, we cropped the whole hand radiographs based on the official bounding boxes and keypoints \cite{ref10}. Then we resized the cropped regions to 500×500 pixels, generating both the cropped images and the attention maps, and applied data augmentation methods including brightness adjustment, contrast adjustment, random affine transformation, and horizontal flipping to the cropped images. The cropped images serve as the input to the Global Feature Extraction Channel, while the attention maps are fed into the Local Feature Extraction Channel. The Global Feature Extraction Channel captures global hand morphology features, whereas the Local Feature Extraction Channel focuses on local epiphyseal regional features. Both types of features are critical in the clinical BAA process. The global and local features are then concatenated along the channel dimension, thereby achieving comprehensive and efficient utilization of multilevel skeletal information. Subsequently, the fused features undergo further extraction and fusion by the Inception-V3 network. Finally, the predicted bone age is obtained through two fully connected layers and a linear layer. In the following parts of this section, we will offer a comprehensive and detailed exposition of each component of the network.
\subsection{Datasets and evaluation indicator}
\begin{figure*}[htbp]
    \centering
    \includegraphics[width=0.8\textwidth, trim=0cm 0cm 0cm 1cm, clip]{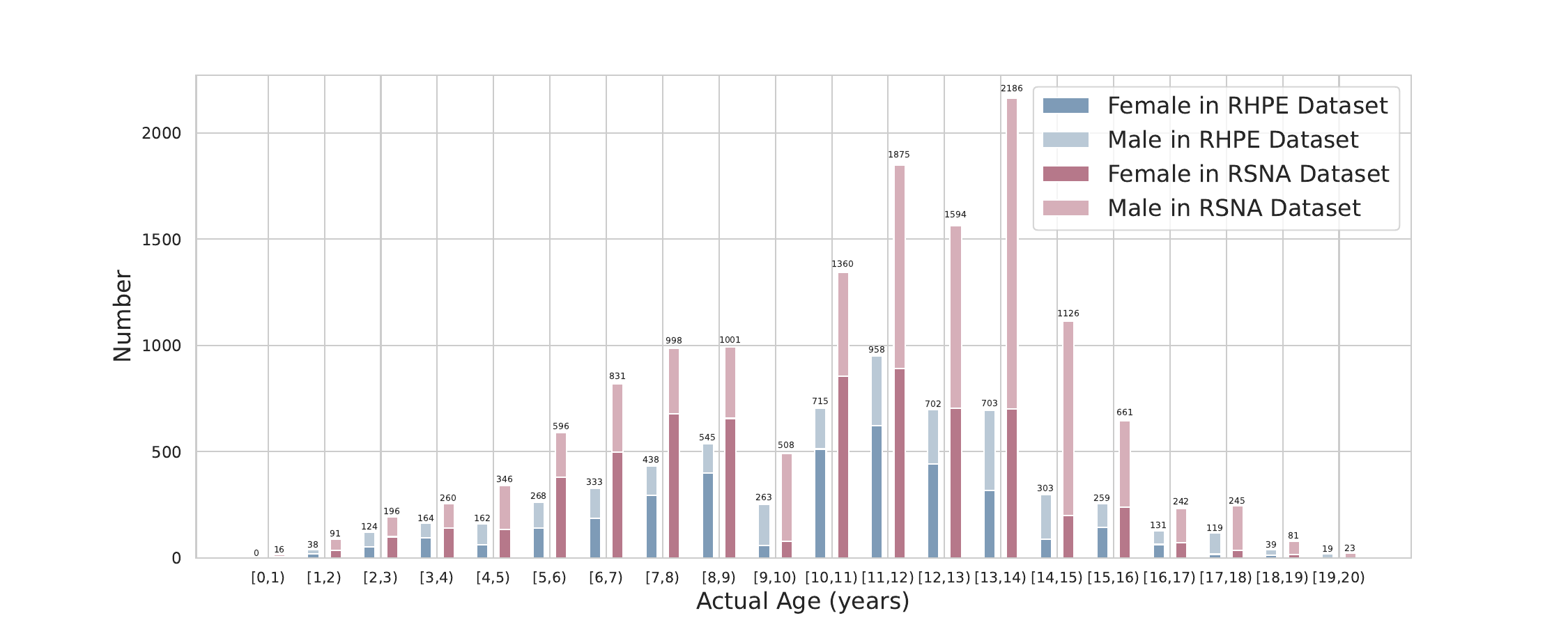}
    \caption{The sample number distribution across different age groups in two public datasets (including  training and validation sets).}
    \label{fig:bone_age_distribution}
\end{figure*}

The proposed method was validated on two public datasets: the RSNA dataset \cite{cicero2017machine} and the RHPE dataset \cite{ref10}. Both datasets provide official splits that have been widely adopted to ensure comparability across different methods, and all comparative methods in this paper use the same splits. The RSNA dataset, sourced from the 2017 RSNA Pediatric Bone Age Challenge, includes 14,236 X-ray images, with bone ages ranging from 0 to 228 months (0 to 19 years). The dataset is divided into a training set (12,611 images), a validation set (1,425 images), and a test set (200 images). The RHPE dataset includes 6,288 X-ray images, with bone ages ranging from 0 to 240 months (0 to 20 years). It is divided into a training set (5,492 images), a validation set (716 images), and a test set (80 images). The distribution of the size of samples across different age groups in these datasets are shown in Fig.\ref{fig:bone_age_distribution}. It is noted that female patients account for 46\% and male patients account for 54\%. Besides, on the RSNA dataset, the age groups of 1-7 , 8-15 , and 16-20 account for 23.4\%, 72.4\%, and 4.2\% respectively. On the RHPE dataset, the corresponding age groups account for 24.3\%, 70.7\%, and 5.0\%. The uneven distribution of samples across age groups, particularly the limited size of samples in certain age groups, poses a considerable challenge for bone age assessment. 

In this paper, the mean absolute error (MAE) and the root mean square error (RMSE) are used as the evaluation metrics to assess the difference between the actual age and the predicted age. MAE and RMSE are defined as:
\begin{equation}
\text{MAE}=\frac{1}{n}\sum_{i = 1}^{n}\vert y_{i}-\hat{y}_{i}\vert
\end{equation}

\begin{equation}
\text{RMSE}=\sqrt{\frac{1}{n}\sum_{i = 1}^{n}(y_{i}-\hat{y}_{i})^{2}}
\end{equation}
Here \(y_{i}\) is the true bone age, \(\hat{y}_{i}\) is the predicted bone age, and \(n\) is the number of images.
\subsection{Global feature extraction based on Transformer}
In the global feature extraction channel, the cropped image serves as the input. Subsequently, we employ five convolutional layers and two max pooling layers for preliminary feature extraction. This step serves as the foundation for more profound feature exploration. To enable the model to more effectively capture the overall skeletal development in the whole hand radiographs, which is essential for accurate bone age assessment, we introduce a Transformer module for this task. The main goal of integrating this module is to enhance the model's ability to extract global information features, which is directly related to accurately assessing the overall structure and developmental stage of hand bones. 
\begin{figure}[h]
\centering
\includegraphics[width=0.7\textwidth]{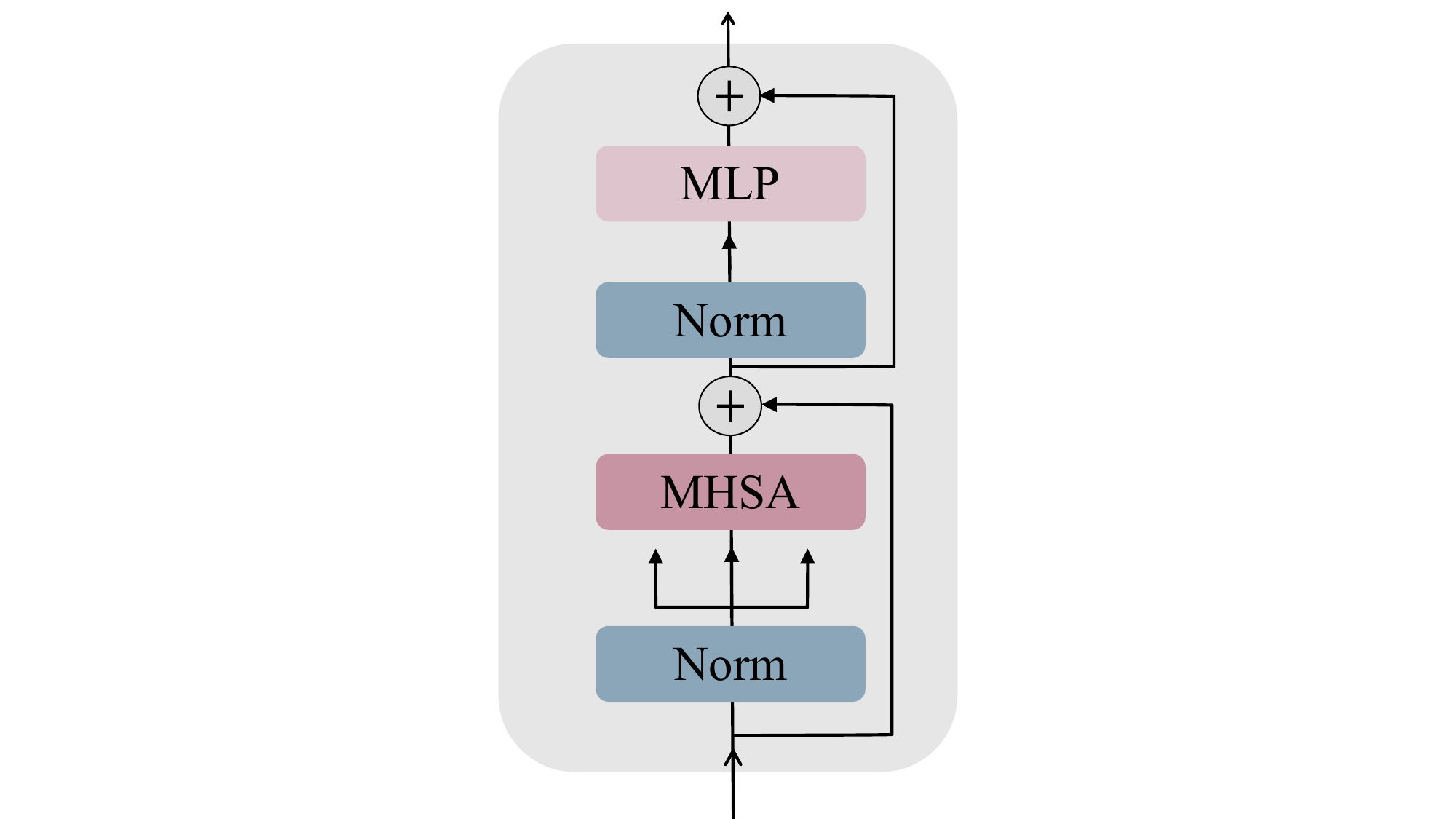}
\caption{The Transformer module \cite{ref12}.}\label{fig:Transformer}
\end{figure}

The Transformer module \cite{ref12} comprises a multi-head self-attention (MHSA) mechanism and a MLP, with residual connections and layer normalization linking them (as depicted in Fig. \ref{fig:Transformer}). The MHSA mechanism is the key advantage of the Transformer module. It consists of multiple independent self-attention heads. Each attention head maps the Query, Key, and Value to multiple different subspaces through different linear transformation methods, then calculates self-attention on each subspace respectively. Finally, the attention outputs obtained from these independent calculations are concatenated, and the output is obtained through a final linear transformation. Specifically, in each attention head, for each position in the input sequence, the model calculates its similarity with all other positions in the sequence, and then updates the representation of this position by weighted averaging the representations of other positions based on these similarities. The specific calculation formula is as follows: 
\begin{equation}
\text{Attention}(Q, K, V) = \text{softmax}\left( \frac{Q K^T}{\sqrt{d_k}} \right) V
\end{equation}

Where \( Q \), \( K \), and \( V \) are the query, key, and value matrices corresponding to the input sequence, and dk is the number of columns of matrices Q and K. The MLP layer is positioned after the MHSA mechanism. It typically consists of two fully connected layers interspersed with a non-linear activation function. Each position in the input sequence is processed independently through these two layers, with the non-linear activation function applied in between. The primary purpose of this structure is to first project the data into a higher-dimensional space and then map it back to a lower-dimensional space, thereby enabling more in-depth feature extraction. Mathematically, given an input X, the output of the MLP layer can be expressed as: 
\begin{equation}
\operatorname{FFN}(x)=\max \left(0, x W_1 + b_1\right) W_2 + b_2
\end{equation}
Where \(W_1\) and \(W_2\) are weight matrices, \(b_1\) and \(b_2\) are bias vectors, and \(\max(0, \cdot)\) denotes the ReLU activation function. This two-step transformation allows the model to capture complex non-linear relationships in the feature space, enhancing its representational capacity while maintaining computational efficiency. 

\subsection{Local feature extraction based on RFAConv}
\begin{figure*}[htbp]
    \centering
    \includegraphics[width=0.8\textwidth, trim=0cm 0cm 0cm 2cm, clip]{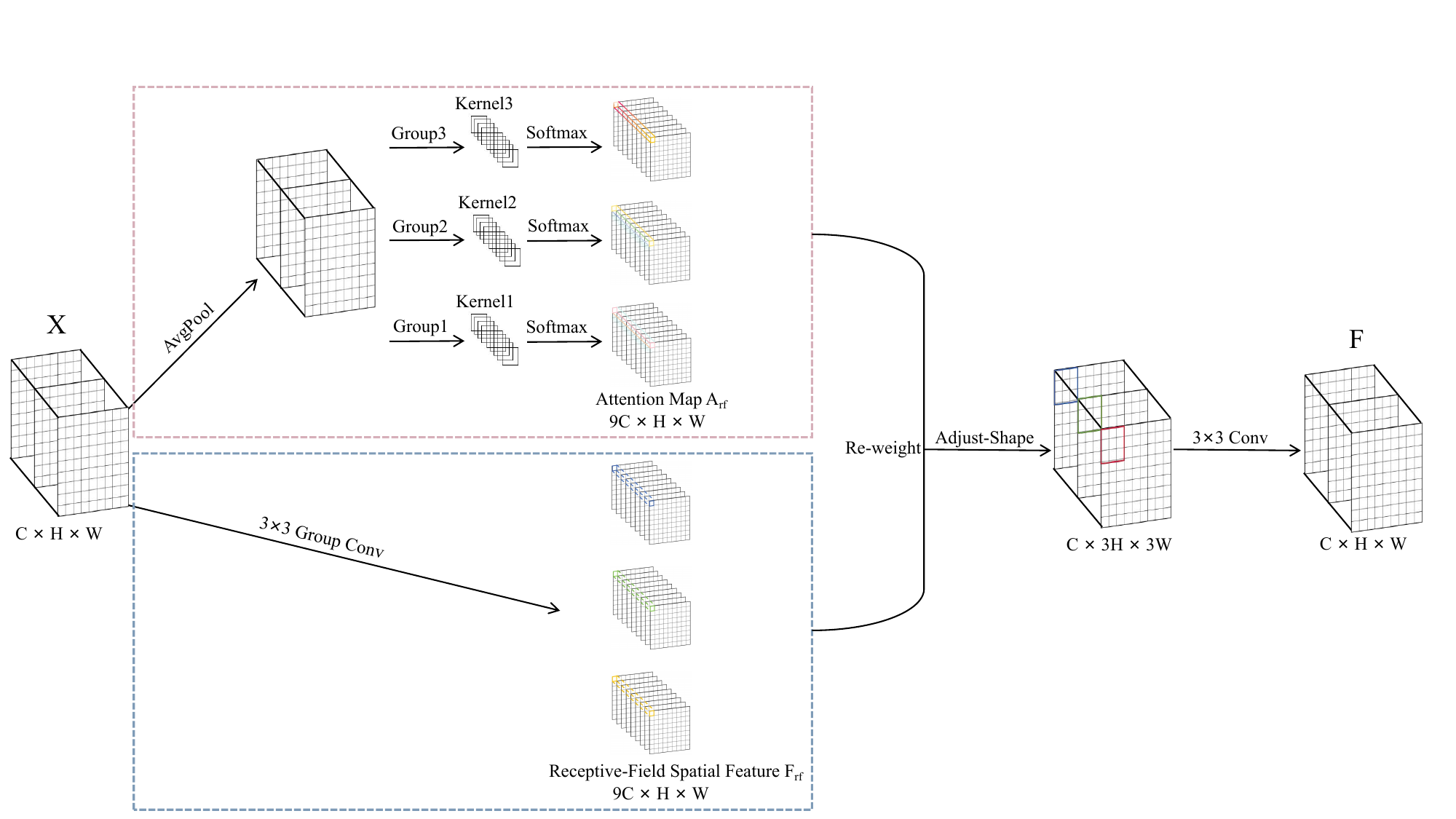}
    \caption{The RFAConv module \cite{ref11}.}
    \label{fig:RFAConv}
\end{figure*}

In the local feature extraction channel, the attention map serves as the input. Similar to the global feature extraction channel, the local feature extraction channel initially utilizes five convolutional layers and two max pooling layers for preliminary feature extraction. However, considering the limited scope of key point regions and their incapability to cover all crucial skeletal information comprehensively, we introduce a RFAConv module \cite{ref11} (as shown in Fig.\ref{fig:RFAConv}) at the end of the local feature extraction channel. The output feature map of the second max pooling layer serve as the input feature map $X$  for RFAConv. RFAConv employs grouped convolutions to extract the receptive field spatial feature $F_{rf}$, and generates attention map $A_{rf}$ through average pooling and $1 \times 1$ grouped convolution operations, which are then normalized using the \text{Softmax} function. The output feature map $F$ is obtained by multiplying the attention map $A_{rf}$ with the transformed receptive field spatial feature $F_{rf}$. The computational expression is as follows:

\begin{equation}
\begin{aligned}
F &= A_{rf} \times F_{rf} \\
  &= \text{Softmax}(g^{1 \times 1}(\text{AvgPool}(X))) \times \text{ReLU}(\text{Norm}(g^{3 \times 3}(X)))
\end{aligned}
\end{equation}

The receptive-field attention can generate independent attention maps for each receptive field feature\cite{ref11}. This effectively overcomes the problem of shared attention maps within the receptive field, which exists in traditional methods. The design of such independent attention maps allows RFAConv to more flexibly emphasize the importance of different features within the receptive field. As a result, it can precisely capture more crucial regions and a greater amount of feature information, thus presenting a more comprehensive and detailed feature representation for the subsequent bone age assessment process. 

\subsection{Feature fusion and boneage prediction}

After extracting global and local features, we concatenate the features along the channel dimension and fuse the features using the Inception-V3 network. Then we incorporate gender information during the prediction stage to enhance the model's expressiveness. By taking into account the significant impact of gender on skeletal development, this improvement helps to enhance the accuracy of BAA. Finally, the predicted bone age is obtained through two fully connected layers and a linear layer. We use smooth L1 loss \cite{ref25} as the loss function to penalize the difference between the prediction ages and the true labels. The computational expression is as follows:

\begin{equation}
\label{eqn:smooth_l1}
\text{Smooth L1 loss} = \begin{cases} 
\frac{1}{2}(y - \hat{y})^2 & \text{if} |y - \hat{y}| < 1 \\
|y - \hat{y}| - \frac{1}{2} & \text{otherwise}
\end{cases}
\end{equation}
Here, \( y - \hat{y} \) represents the difference between the actual age $y$ and the predicted age $\hat{y}$. When \( |y - \hat{y}| \) is less than \( 1 \), it employs the quadratic term of L2 loss to avoid the constant gradient problem of L1 loss near zero and can achieve quick convergence. When \( |y - \hat{y}| \) is greater than or equal to 1, it switches to the linear term of L1 loss to limit the gradient and enhance robustness to outliers. This design makes the Smooth L1 loss function more robust to outliers than the L2 loss, and smoother than the L1 loss when the loss value is small. It not only stabilizes the training process but also optimizes the overall performance of the model. Therefore, we use the Smooth L1 loss function during the training process to improve the model's performance and stability.

\section{Experiments}\label{sec:experiments}
\subsection{Experimental details}

We implemented the proposed method in Pytorch and trained on a NVIDIA GeForce RTX 4090 GPU for a total of 150 epochs. The initial learning rate was set to 0.001, with a batch size of 17 images per batch, and the Adam optimizer \cite{kingma2014adam} with its standard parameters was used for model optimization. Additionally, a dynamic learning rate scheduler and early stopping mechanism were employed. A dynamic learning rate scheduler was used to reduce the learning rate when reaching a plateau with a patience of 2 epochs, a reducing factor of 0.8, and a cooldown of 5 epochs.

\subsection{Effectiveness Experiments}
\begin{figure*}[!ht]
    \centering
    \subfloat[]{\includegraphics[width=0.44\textwidth]{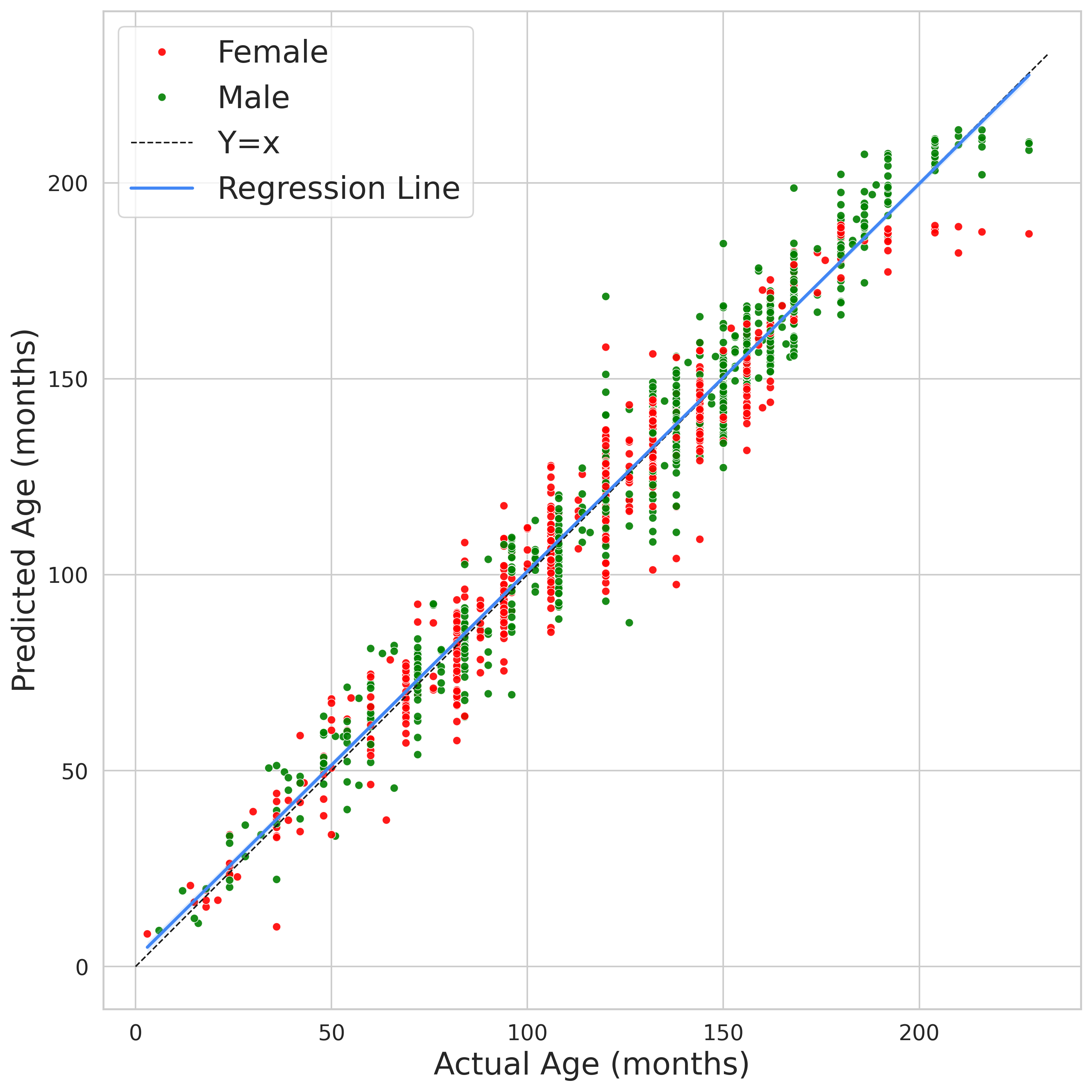}}
    \hspace{0.1\textwidth}
    \subfloat[]{\includegraphics[width=0.44\textwidth]{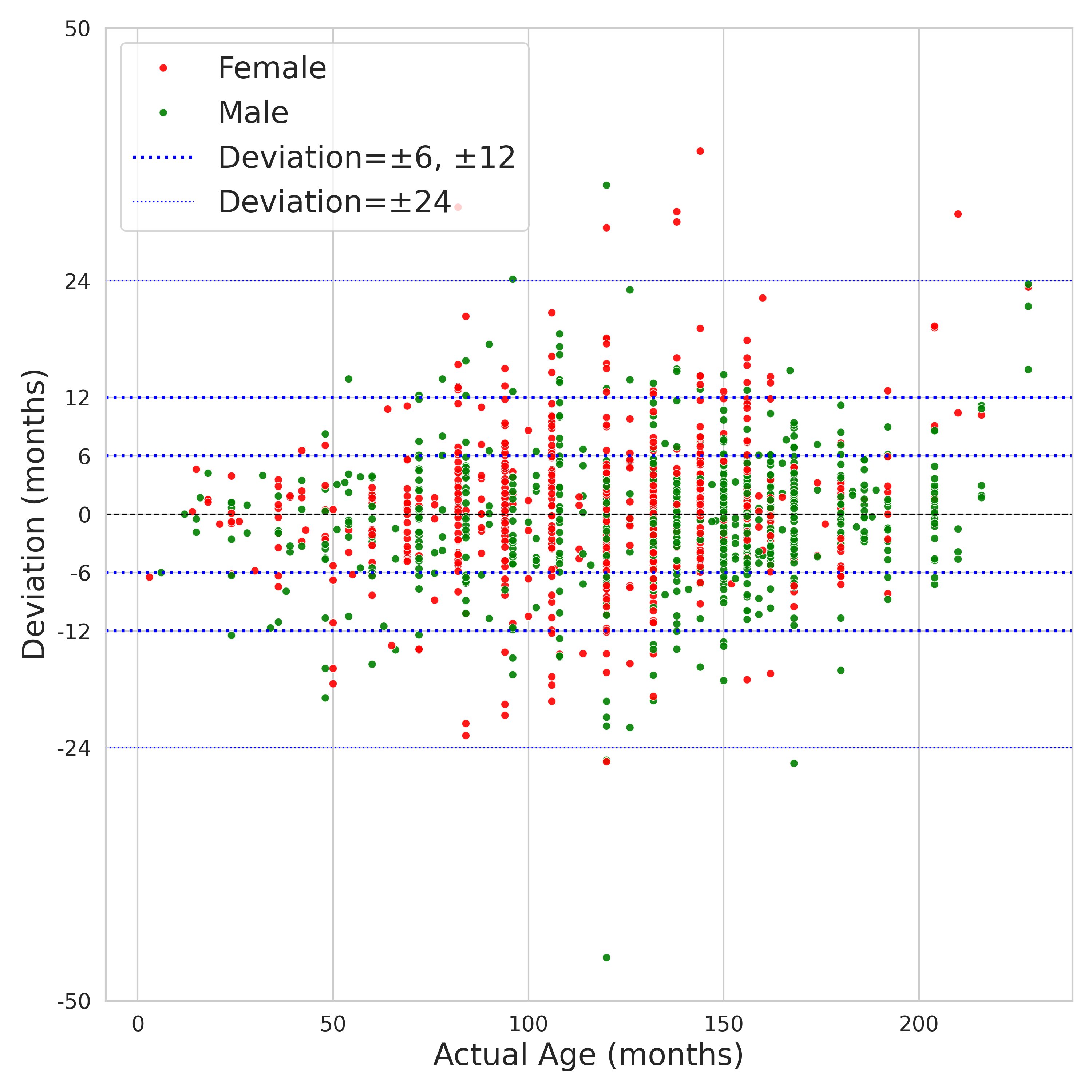}}
    \\[0.5ex]
    \subfloat[]{\includegraphics[width=0.44\textwidth]{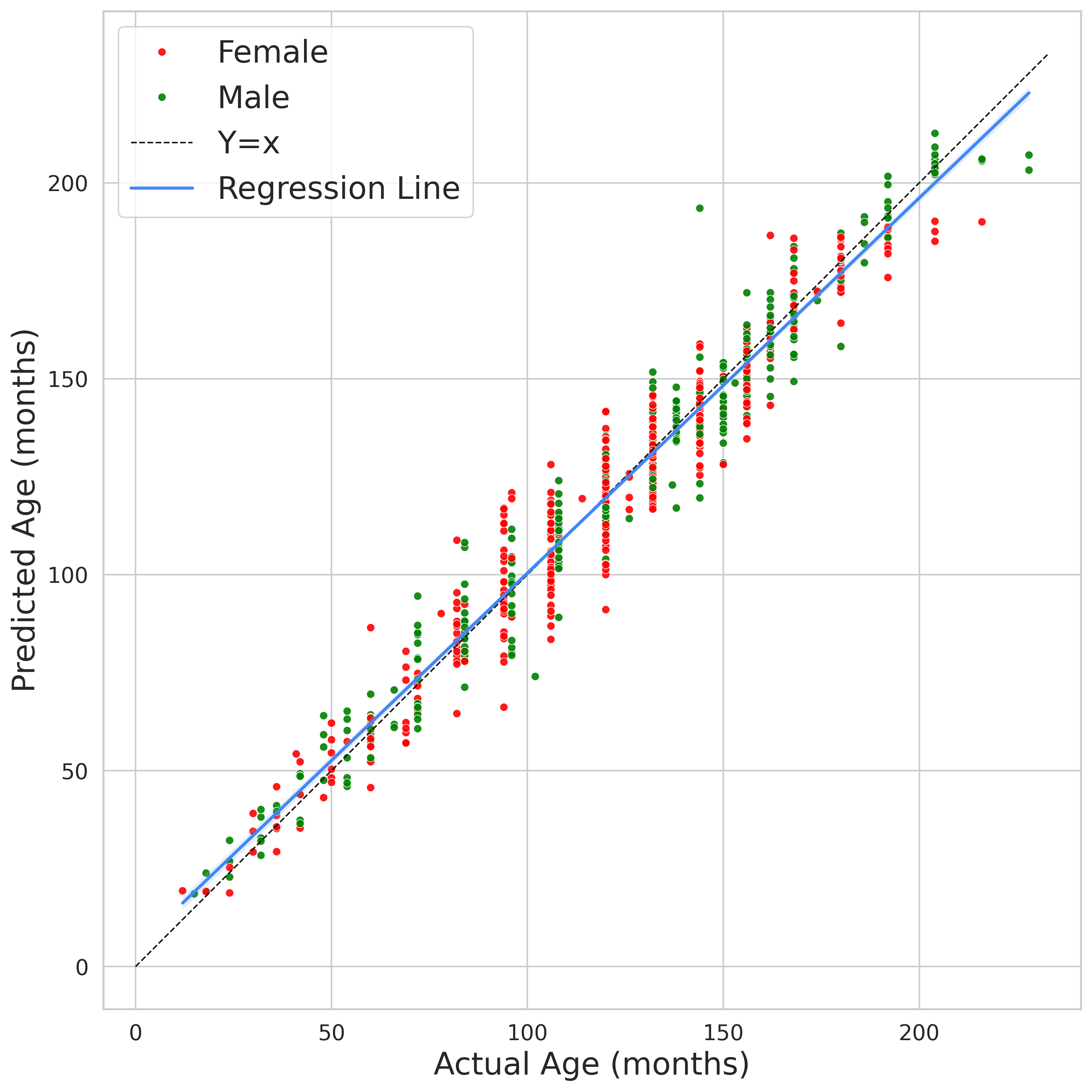}}
    \hspace{0.1\textwidth}
    \subfloat[]{\includegraphics[width=0.44\textwidth]{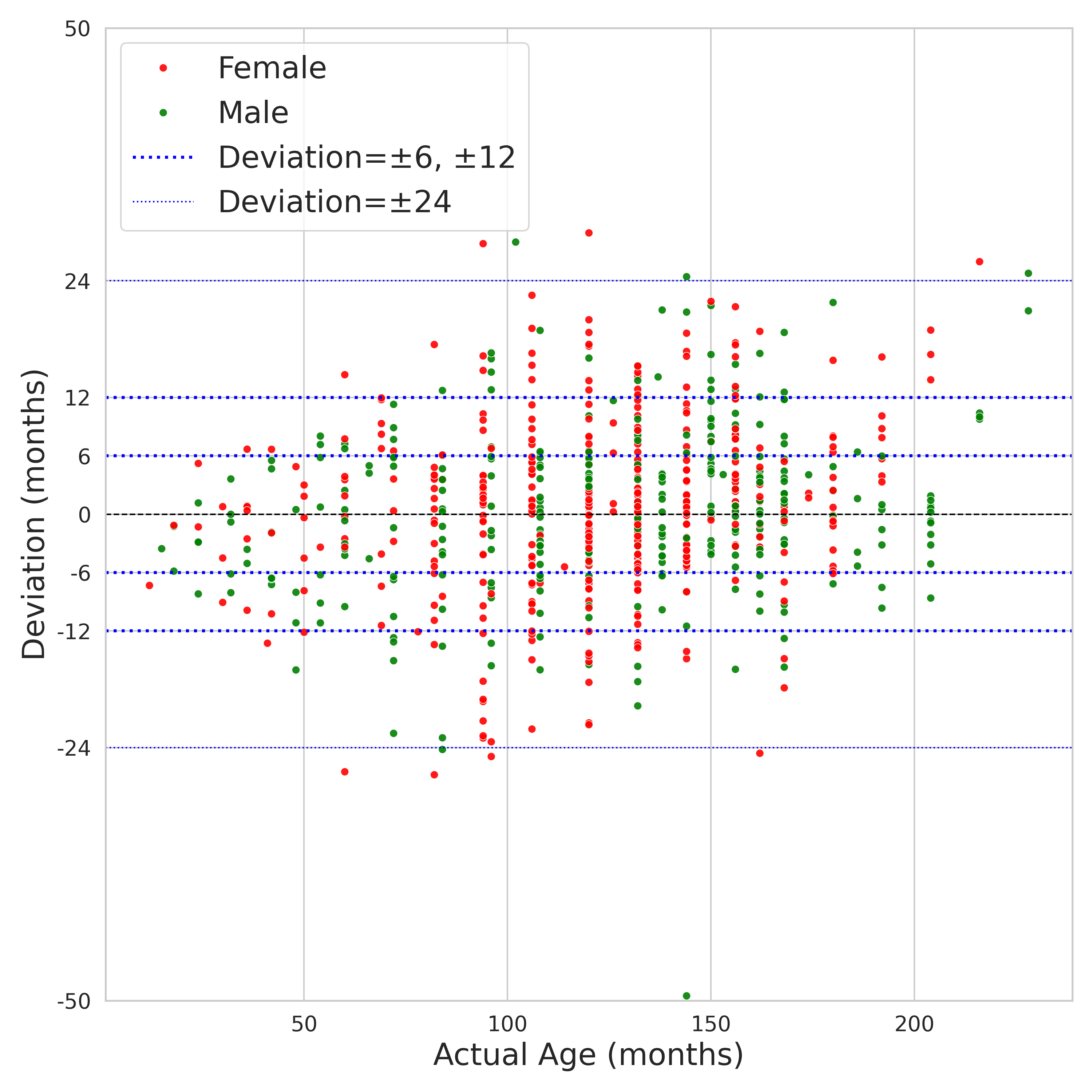}}
    \caption{Statistical results of the proposed method in bone age assessment. (a) Actual age and predicted age on the RSNA validation dataset. (b) Actual age and deviation on the RSNA validation dataset. (c) Actual age and predicted age on the RHPE validation dataset. (d) Actual age and deviation on the RHPE validation dataset.}
    \label{fig:fpur}
\end{figure*}

To validate the effectiveness of the proposed method, we conduct training and testing on both the RSNA and RHPE datasets. The method achieves an MAE of 3.81 months on the RSNA test set and 5.65 months on the RHPE test set (and an RMSE of 5.08 months on the RSNA test set). The Pearson correlation significance test shows p-values of 1.4e-10 (on the RSNA test set) and 1.6e-11 (on the RHPE test set), confirming a highly positive correlation between predicted and actual ages. To further explore the prediction results across different age groups, given that the validation sets have a larger volume, we present the bone age prediction and deviation graphs on the two validation datasets (as shown in Fig.\ref{fig:fpur}). The method achieves MAE of 4.88 months on the RSNA validation set and 6.74 months on the RHPE validation set (The method achieves RMSE of 6.96 months on the RSNA validation set and 9.02 months on the RHPE validation set), a p-value of 1.0e-8 for the RSNA validation set further confirms the model’s overall significance.

Fig.\ref{fig:fpur}(a) and (c) display the distribution of the actual age and the predicted age, there is a obvious consistency between the predicted age and the actual age calibrated by doctors in most cases. Fig.\ref{fig:fpur}(b) and (d) display the distribution of prediction deviations, with most deviations within 12 months. On the RSNA validation dataset, the cumulative accuracies within deviations of 6 months and 12 months are 72.8\% and 91.3\% respectively. For females, these accuracies are 70\% and 89.8\%, and for males, they are 75.1\% and 92.6\%. On the RHPE validation dataset, the cumulative accuracies within deviations of 6 months and 12 months are 56.9\% and 82.6\% respectively. For females, these accuracies are 56\% and 81\%, and for males, they are 58\% and 84\%. The better prediction results for males were likely due to their higher proportion on the datasets. The RSNA dataset has better prediction results than the RHPE dataset because of its larger volume, which benefits the prediction model. 

When analyzing the data of different age groups, we find that the limited sample size in the 16-20 age group directly leads to severely hampered prediction performance of the model. On the RSNA validation dataset, the cumulative accuracies within deviations of 6 months and 12 months are 62.9\% and 87\% respectively. On the RHPE validation dataset, the corresponding accuracy rates are 52.9\% and 79.4\% respectively. In the 8-15 age group, despite a relatively large size of samples, the complexity of bone age assessment is extremely high. During puberty within this age range, bones develop rapidly. This development is accompanied by intricate and subtle morphological changes in bones, with some regions showing overlapping features. Adding to this complexity, there are noticeable disparities in the pubertal development speed among individuals, which further compounds the challenge of accurately predicting bone age. On the RSNA validation dataset, the cumulative accuracies within deviations of 6 months and 12 months are 73.5\% and 91.6\% respectively. On the RHPE validation dataset, the corresponding accuracy rates are 58\% and 82.1\% respectively. Compared with the 16-20 age group, the prediction accuracy rates within the same deviation intervals on the RSNA validation dataset show an improvement. The primary reason is the limited sample size in the 16-20 age group, which greatly restricts the model's ability to obtain sufficient representative information. Additionally, the degree of hand bone joint closure in the 16-20 age group is higher than that in the 8-15 age group, resulting in low differentiation between radiographs. In the 1-7 age group, most bones are not fully developed, with large gaps between them. In addition, the strong regularity and consistent changes in the skeletal development of infants and toddlers provide clear features for the model. On the RSNA validation dataset, the cumulative accuracies within deviations of 6 months and 12 months are 72.5\% and 91.2\% respectively. On the RHPE validation dataset, the corresponding accuracy rates are 54.5\% and 84.4\% respectively. Although the sample size of this age group is smaller than that of the 8-15 age group, the overall prediction accuracy rates are similar, further indicating the lower complexity of bone age assessment in this age group. In summary, these results fully demonstrate that the model can provide relatively accurate bone age estimates in most cases. Although there are certain prediction deviations in specific age groups, the overall performance of the model is robust and can provide valuable reference for clinical applications. 

\subsection{Comparison Experiments}
We summarize and compare the MAE values of bone age assessment methods reported in existing literature based on the RSNA dataset, with specific results shown in Table \ref{table:mae_comparison}, all comparative methods in this paper use the same official splits. Compared to the methods that can assess bone age in a single step (e.g. Lee et al. \cite{ref7} and Tang et al. \cite{ref18}), our method comprehensively extract both global and local features. In contrast to the ROI-based methods (e.g. Ji et al. \cite{ref23}, Li et al. \cite{ref24}, and Koitka et al. \cite{ref22}), our approach avoids over-focusing on ROIs and can strike a balance between global and local information. Compared with the attention-learning-based on methods (e.g. Ren et al. \cite{ref26}, Escobar et al. \cite{ref10}, Liu et al. \cite{ref30}, Wu et al. \cite{ref31}, and Yang et al. \cite{ref32}), the RFAConv module in our method effectively compensates for the insufficient coverage of keypoints in attention maps, enabling more comprehensive coverage of key bones. By fusing global and local features, our method achieves MAE of 3.81 months, outperforming existing methods.
\begin{table}[htbp]
\caption{Comparison of different SOTA methods on the RSNA dataset.}
\label{table:mae_comparison}
\centering
\begin{tabular}{lc}
\toprule
\textbf{Methods} & \textbf{MAE (months)} \\
\midrule
Lee et al. \cite{ref7} & 18.9 \\
Tang et al. \cite{ref18} & 5.53 \\
Ren et al. \cite{ref26} & 5.2 \\
Koitka et al. \cite{ref22} & 4.56 \\
Ji et al. \cite{ref23} & 4.49 \\
Escobar et al. \cite{ref10} & 4.14 \\
Liu et al. \cite{ref30} & 4.10 \\
Li et al. \cite{ref24} & 4.09 \\
Wu et al. \cite{ref31} & 3.91 \\
Yang et al. \cite{ref32} & 3.88 \\
Ours & \textbf{3.81} \\
\bottomrule
\end{tabular}
\end{table}

\begin{table}[htbp]
\caption{Comparison of different SOTA methods on the RHPE dataset.}
\label{table:mae_comparison1}
\centering
\begin{tabular}{lc}
\toprule
\textbf{Methods} & \textbf{MAE (months)} \\
\midrule
Escobar et al. \cite{ref10} & 7.60 \\
Gonz\'{a}lez et al. \cite{ref29} & \textbf{5.47} \\
Li et al. \cite{ref24} & 6.78 \\
Ours & \underline{5.65} \\
\bottomrule
\end{tabular}
\end{table}

Table \ref{table:mae_comparison1} presents a comprehensive summary and comparison of bone age assessment methods reported in the literature using the RHPE dataset. Our method was compared with three methods: Escobar et al. \cite{ref10}, González et al. \cite{ref29} and Li et al. \cite{ref24}. The method proposed by González et al. \cite{ref29} achieved the best MAE of 5.47 months, with patients' chronological ages being used as prior information. Without the chronological age, our method achieves the second best MAE of 5.65 months, only 0.18 months higher than the best result.

\subsection{Ablation Experiments}

To systematically evaluate the distinct impact of the Transformer module and the RFAConv module on model performance, we conduct comprehensive ablation experiments on two datasets. The detailed experimental results are shown in Table \ref{table:mae_comparison2}. On the RSNA dataset, when only the Transformer module is added to the baseline network, the MAE is 4.38 months. Adding only the RFAConv module lead to a MAE of 4.22 months. When both modules are incorporated, the MAE dropped to 3.81 months. This represents a 12.8\% improvement over adding only the Transformer module and a 10.2\% improvement over adding only the RFAConv module. On the RHPE dataset, the baseline network with adding the Transformer module alone has MAE of 6.10 months, with adding the RFAConv module alone has MAE of 6.40 months, and with both added has MAE of 5.65 months. This represents a 7.4\% improvement over adding only the Transformer module and a 12.3\% improvement over adding only the RFAConv module. Both results show that the combination of the RFAConv module and the Transformer module is beneficial for enhancing model performance. 
\begin{table}[!ht]
\centering
\caption{Impact of different modules on model performance.}
\label{table:mae_comparison2}
\begin{tabular}{lccc}
\toprule
\multicolumn{1}{c}{\textbf{Dataset}} & \multicolumn{2}{c}{\textbf{Module}} & \textbf{MAE (months)} \\ 
\cmidrule(lr){2-3}
& \textbf{Transformer} & \textbf{RFAConv} & \\ 
\midrule
RSNA & $\times$ & $\times$ & / \\ 
& $\checkmark$ & $\times$ & 4.38 \\ 
& $\times$ & $\checkmark$ & 4.22 \\ 
& $\checkmark$ & $\checkmark$ & \textbf{3.81} \\ 
\midrule
RHPE & $\times$ & $\times$ & 7.60 \\ 
& $\checkmark$ & $\times$ & 6.10 \\ 
& $\times$ & $\checkmark$ & 6.40 \\ 
& $\checkmark$ & $\checkmark$ & \textbf{5.65} \\ 
\bottomrule
\end{tabular}
\end{table}

\begin{figure}[t]
    \centering
    \setlength{\tabcolsep}{0.5pt}  
    \renewcommand{\arraystretch}{0.2}  
    \begin{tabular}{@{}cccccccc@{}}  
        \includegraphics[width=0.12\textwidth, height=1.8cm, trim=0 0 0 0, clip]{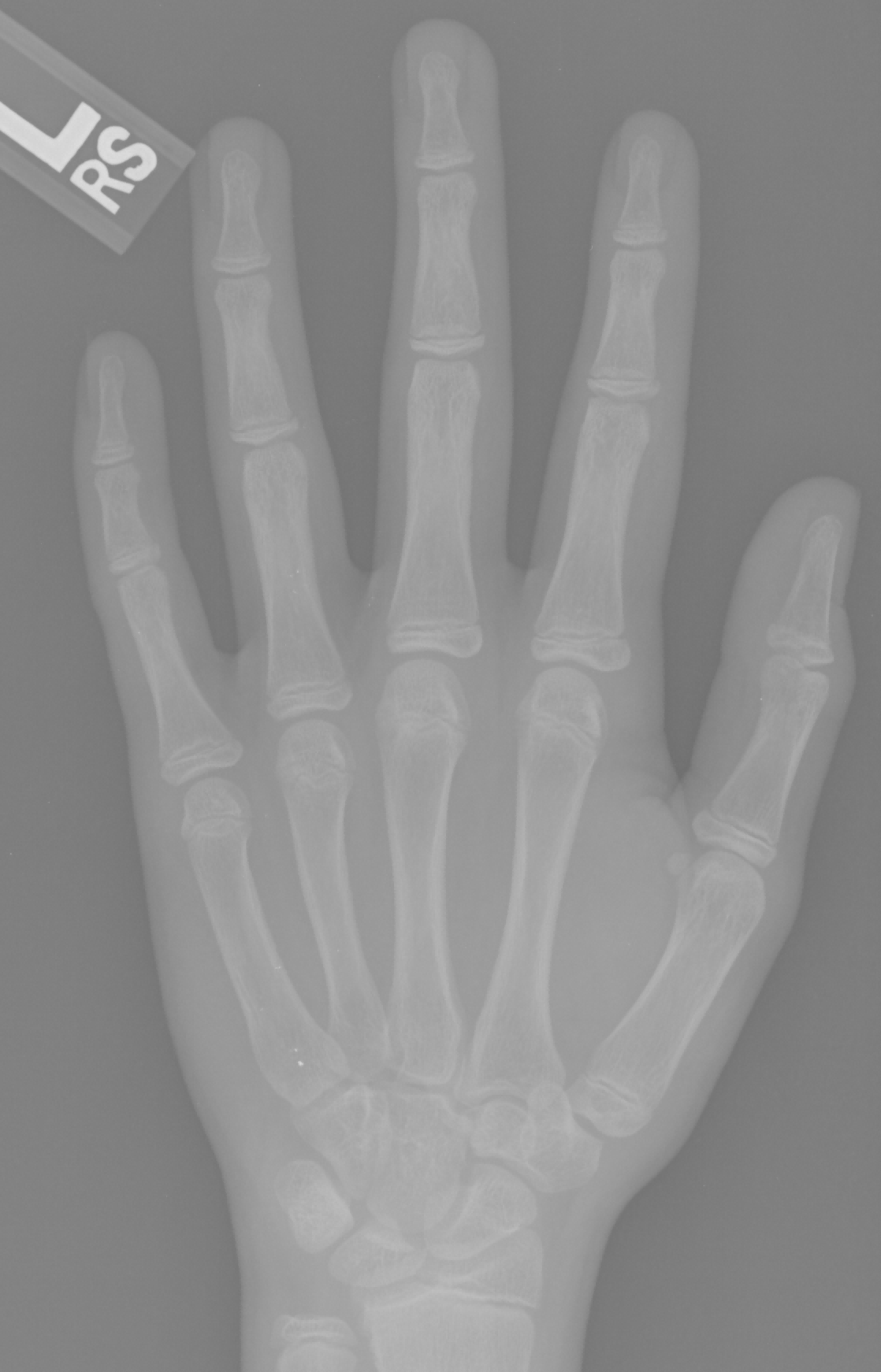} &
        \includegraphics[width=0.12\textwidth, height=1.8cm, trim=0 0 0 0, clip]{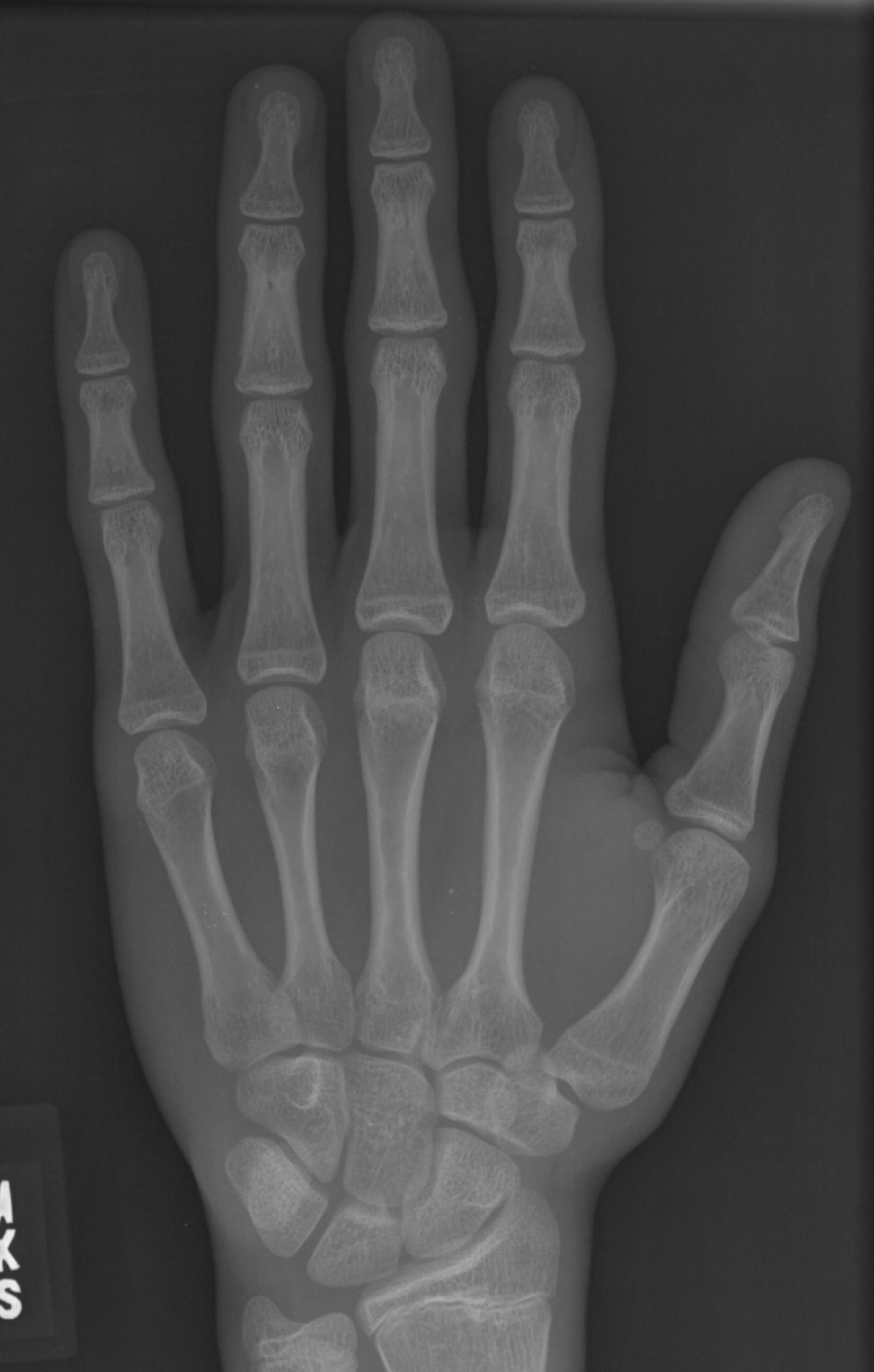} &
        \includegraphics[width=0.12\textwidth, height=1.8cm, trim=0 0 0 0, clip]{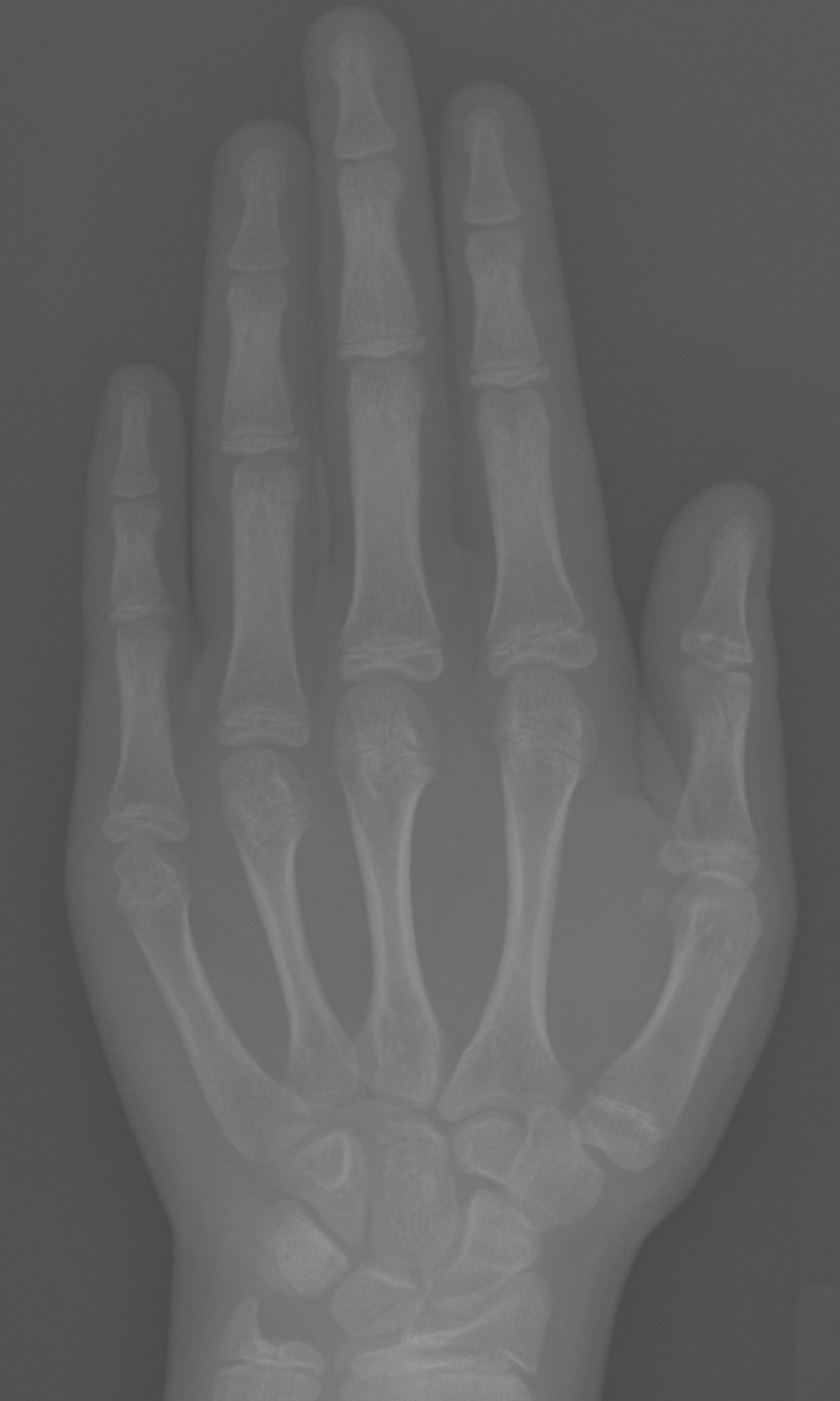} &
        \includegraphics[width=0.12\textwidth, height=1.8cm, trim=0 0 0 0, clip]{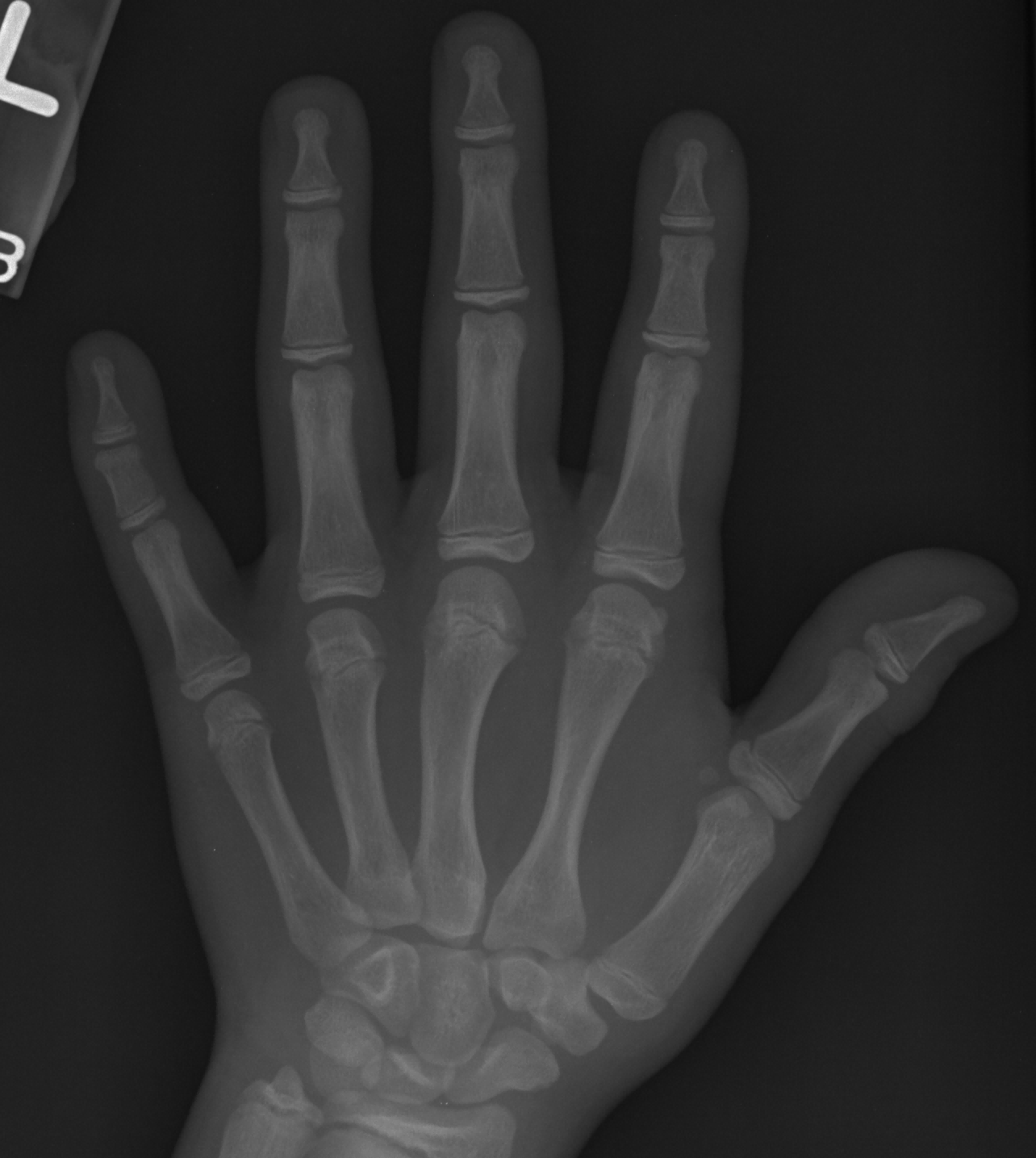} &
        \includegraphics[width=0.12\textwidth, height=1.8cm, trim=0 0 0 0, clip]{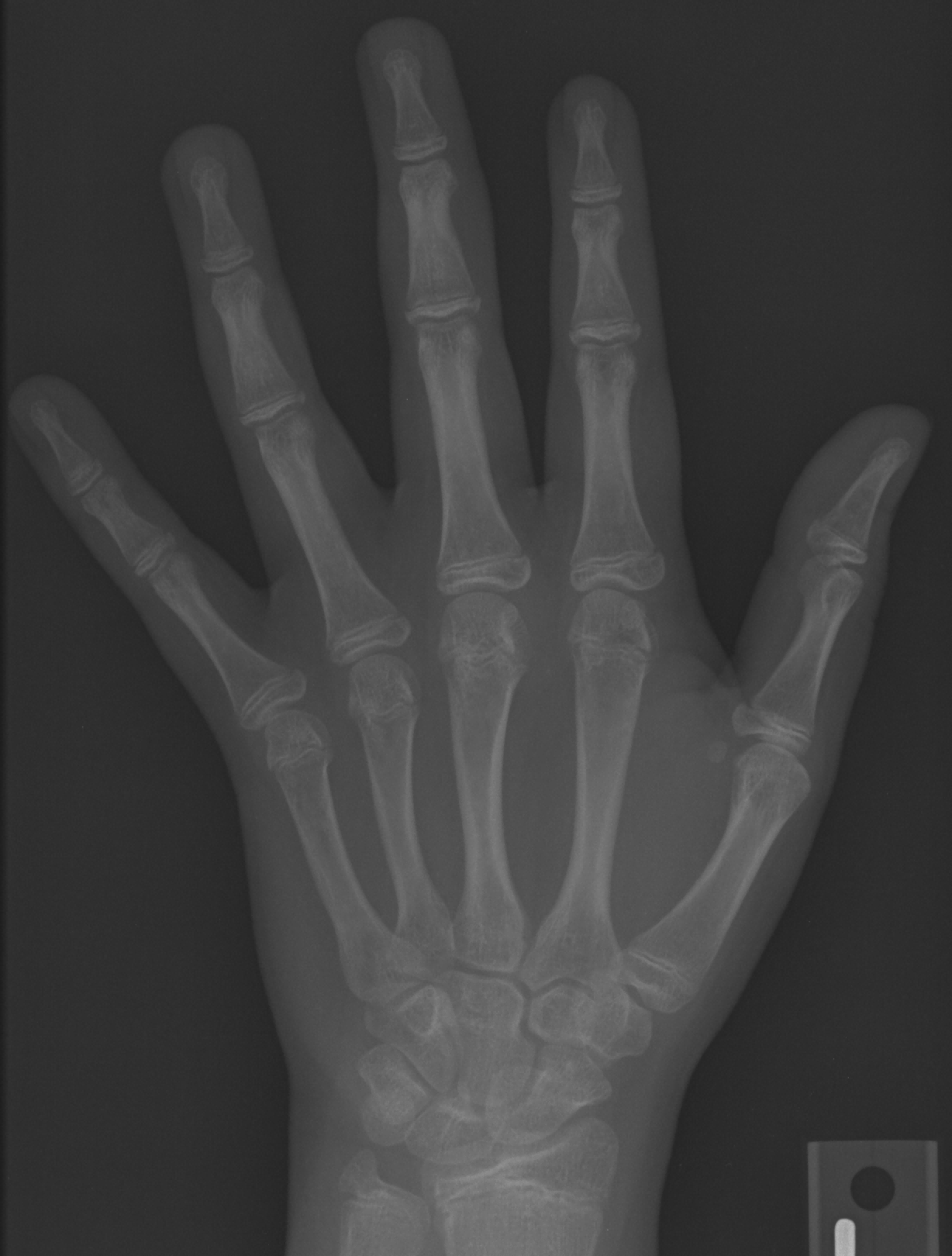} &
        \includegraphics[width=0.12\textwidth, height=1.8cm, trim=0 0 0 0, clip]{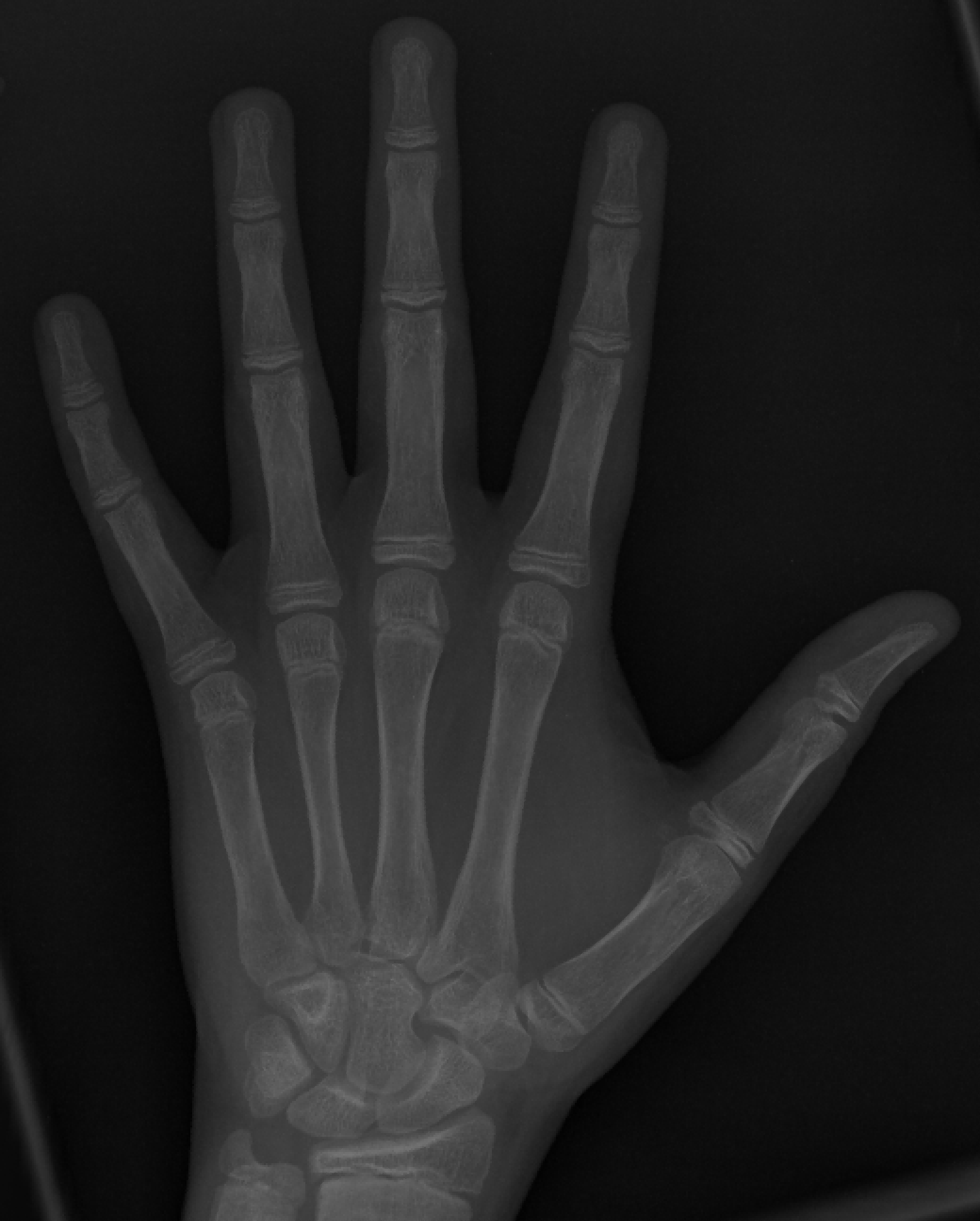} &
        \includegraphics[width=0.12\textwidth, height=1.8cm, trim=0 0 0 0, clip]{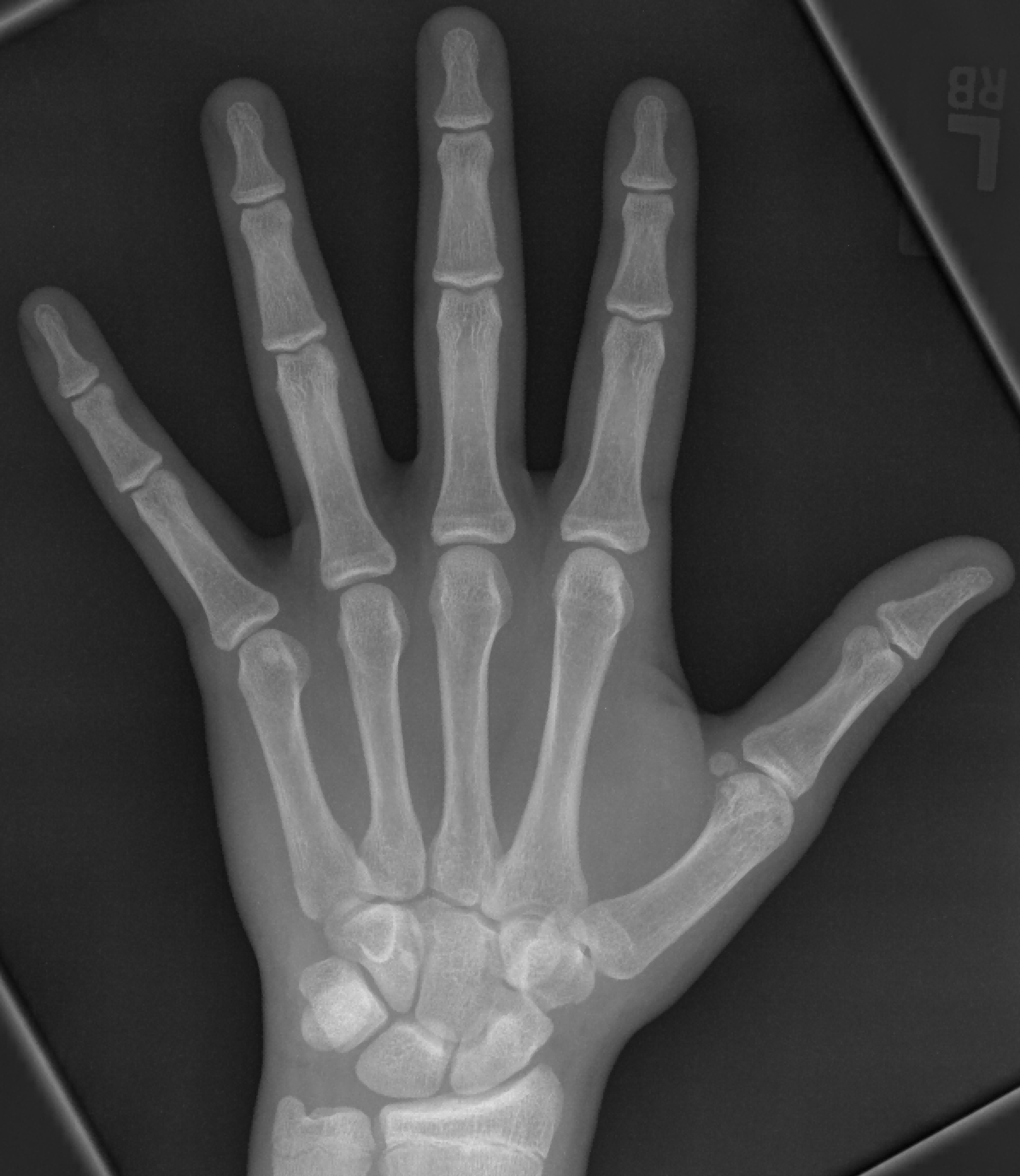} &
        \includegraphics[width=0.12\textwidth, height=1.8cm, trim=0 0 0 0, clip]{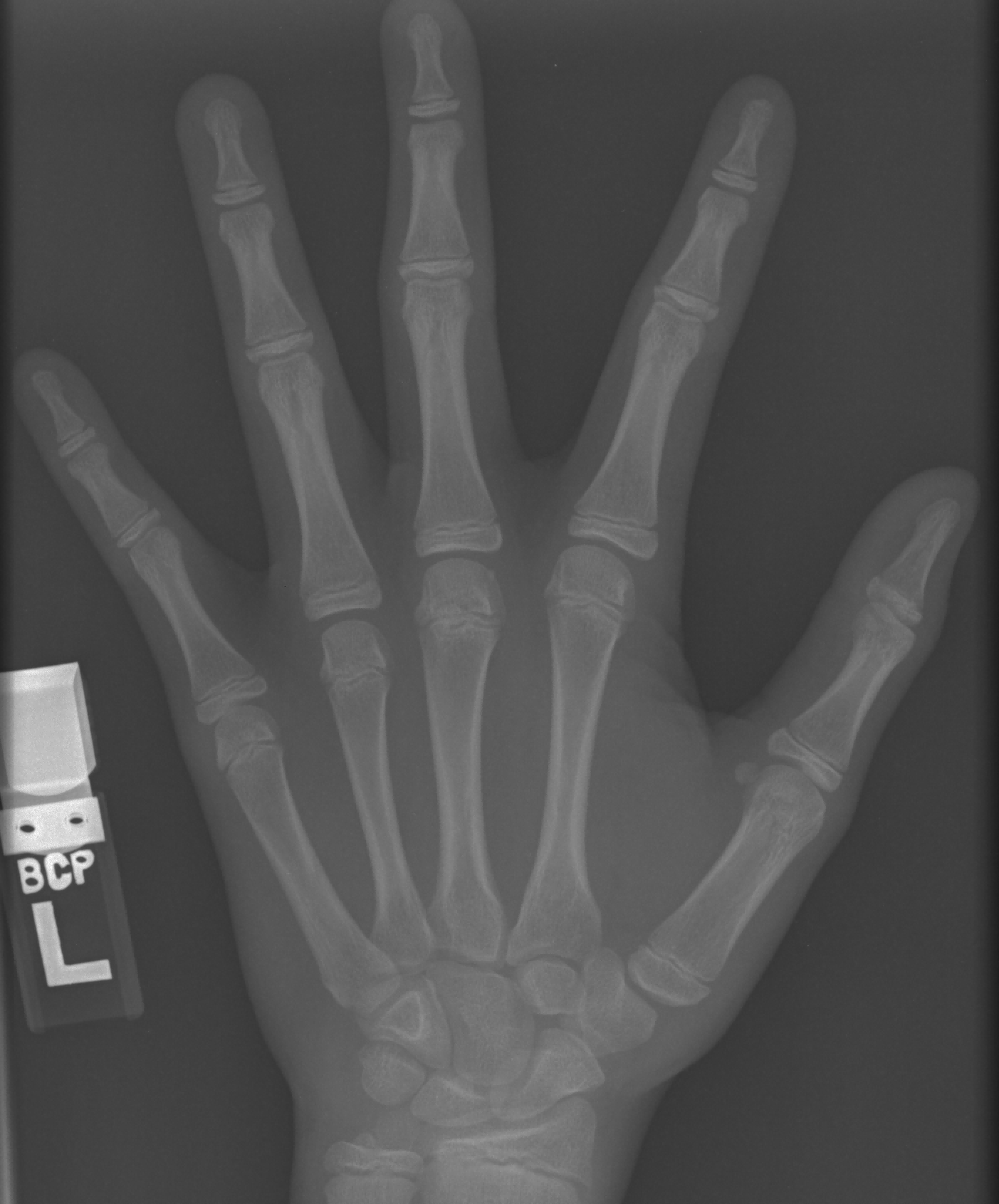} \\
                
        \includegraphics[width=0.125\textwidth, height=2.2cm, trim=0 0 0 0, clip, keepaspectratio]{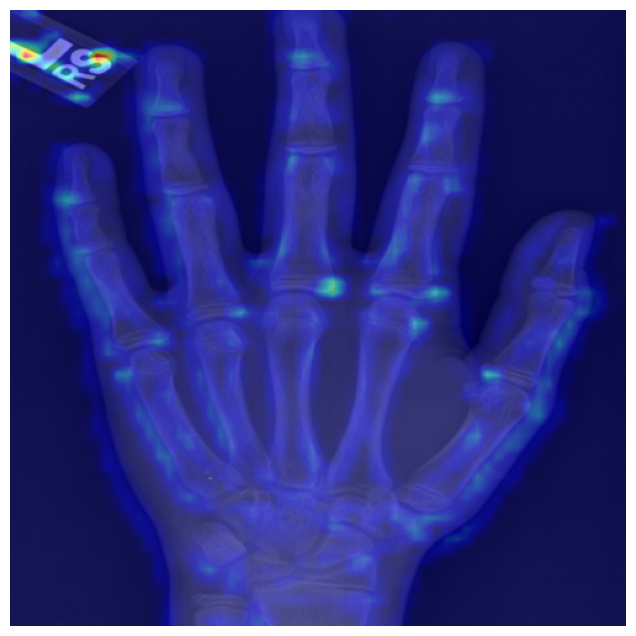} &
        \includegraphics[width=0.125\textwidth, height=2.2cm, trim=0 0 0 0, clip, keepaspectratio]{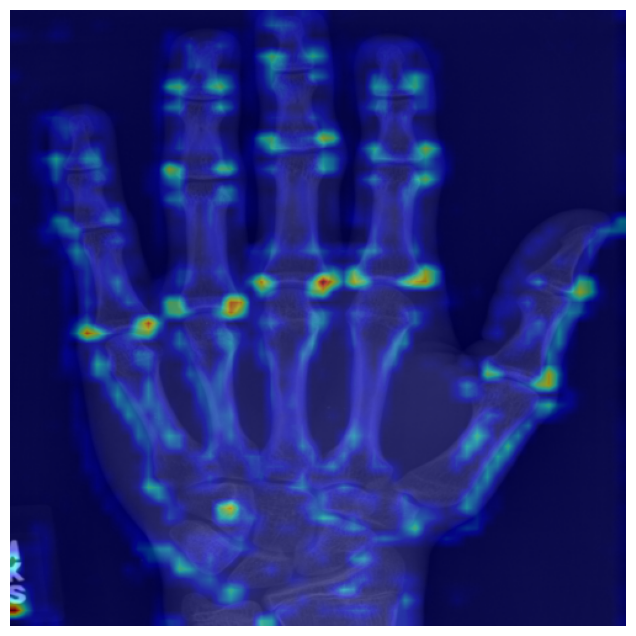} &
        \includegraphics[width=0.125\textwidth, height=2.2cm, trim=0 0 0 0, clip, keepaspectratio]{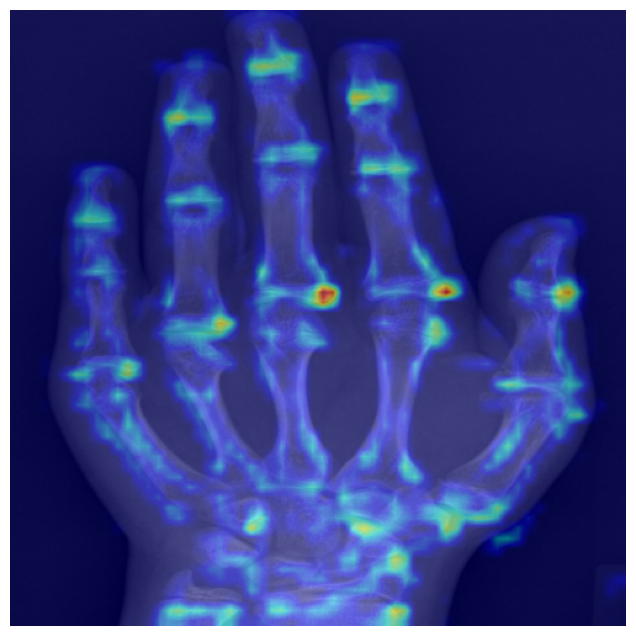} &
        \includegraphics[width=0.125\textwidth, height=2.2cm, trim=0 0 0 0, clip, keepaspectratio]{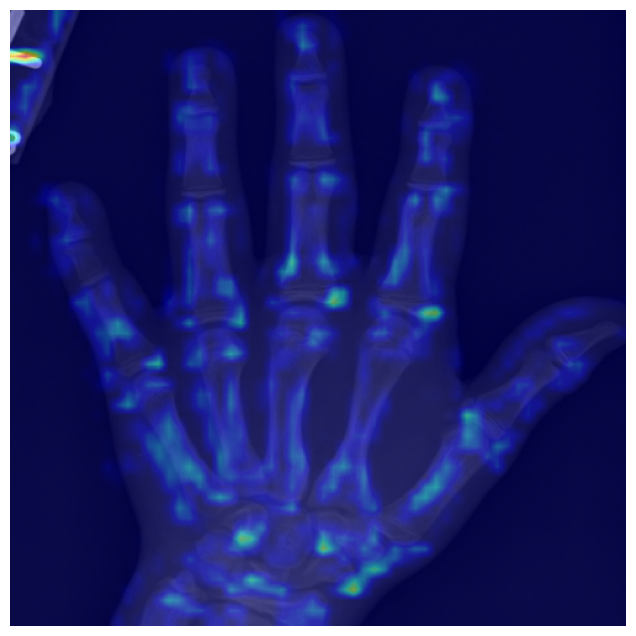} &
        \includegraphics[width=0.125\textwidth, height=2.2cm, trim=0 0 0 0, clip, keepaspectratio]{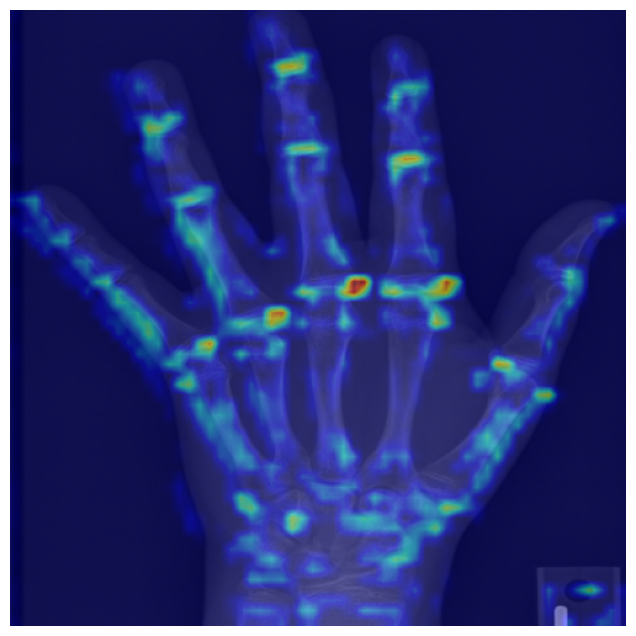} &
        \includegraphics[width=0.125\textwidth, height=2.2cm, trim=0 0 0 0, clip, keepaspectratio]{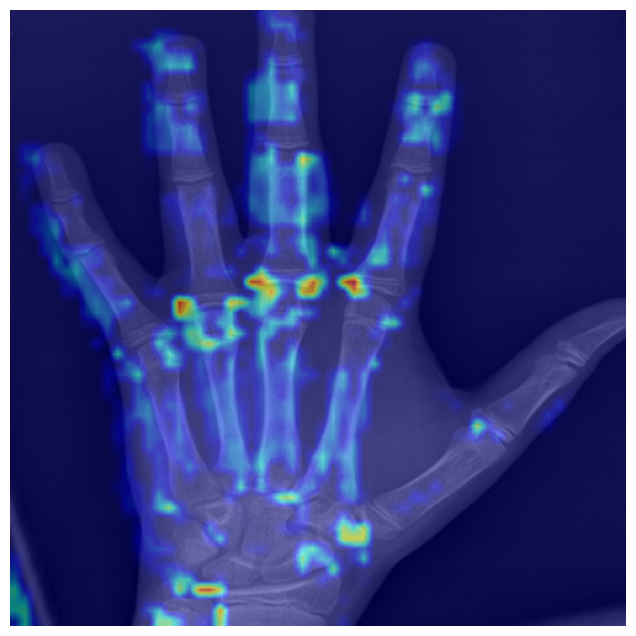} &
        \includegraphics[width=0.125\textwidth, height=2.2cm, trim=0 0 0 0, clip, keepaspectratio]{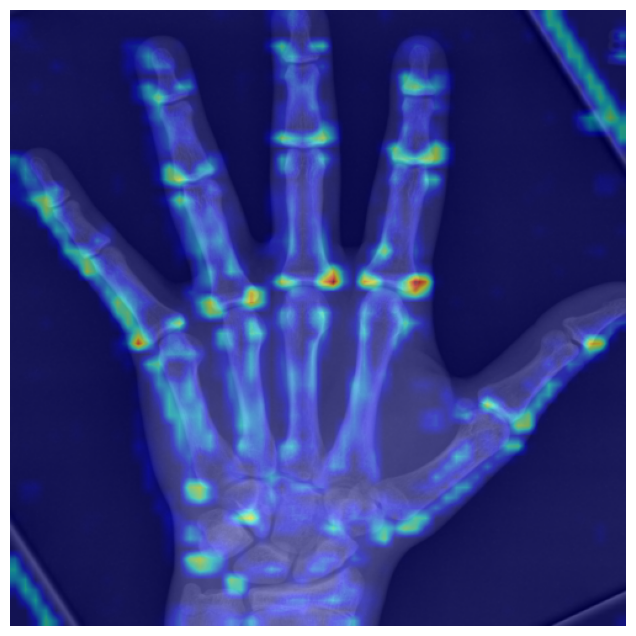} &
        \includegraphics[width=0.125\textwidth, height=2.2cm, trim=0 0 0 0, clip, keepaspectratio]{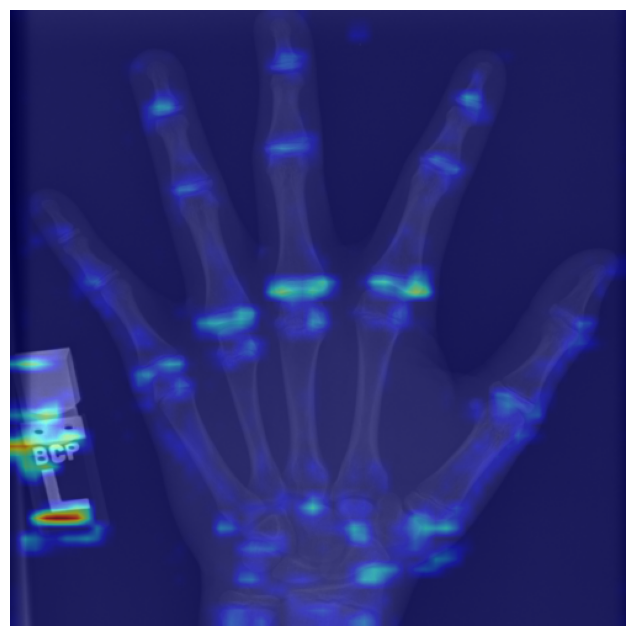} \\
        
        \includegraphics[width=0.125\textwidth, height=2.2cm, trim=0 0 0 0, clip, keepaspectratio]{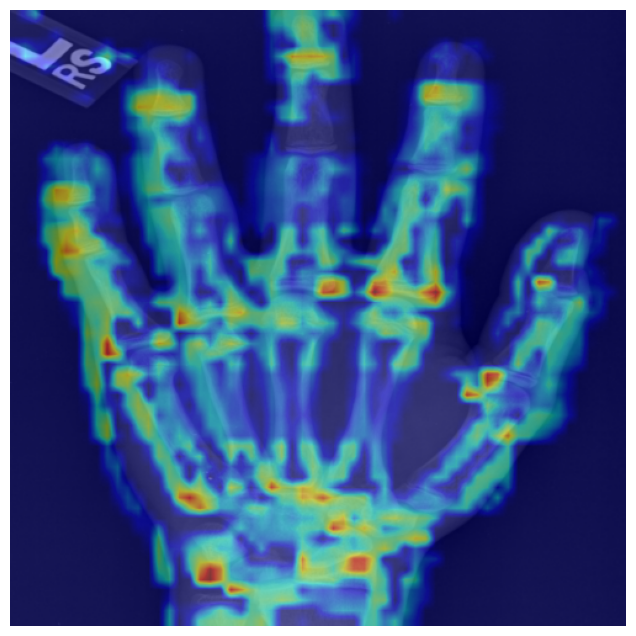} &
        \includegraphics[width=0.125\textwidth, height=2.2cm, trim=0 0 0 0, clip, keepaspectratio]{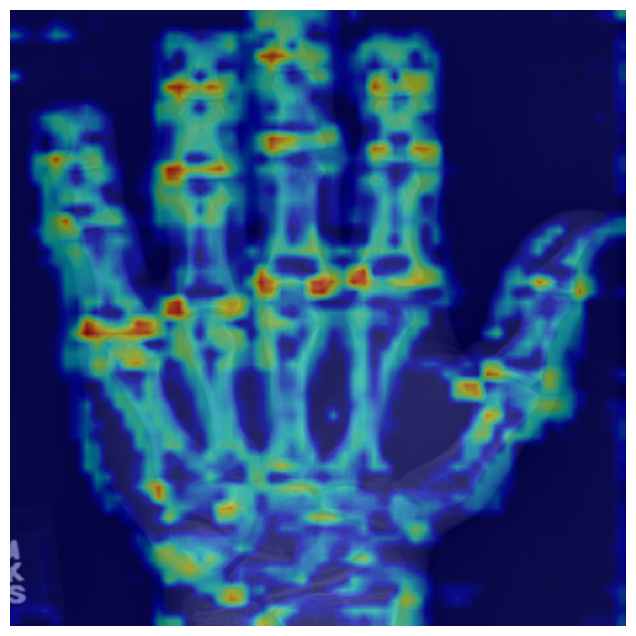} &
        \includegraphics[width=0.125\textwidth, height=2.2cm, trim=0 0 0 0, clip, keepaspectratio]{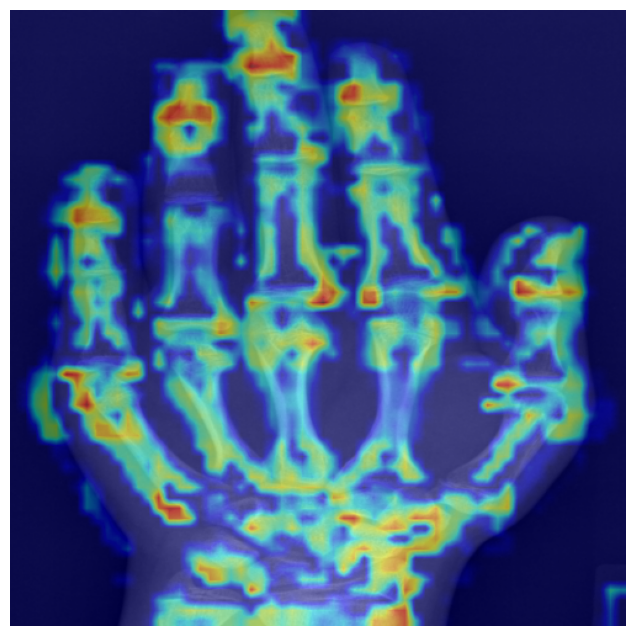} &
        \includegraphics[width=0.125\textwidth, height=2.2cm, trim=0 0 0 0, clip, keepaspectratio]{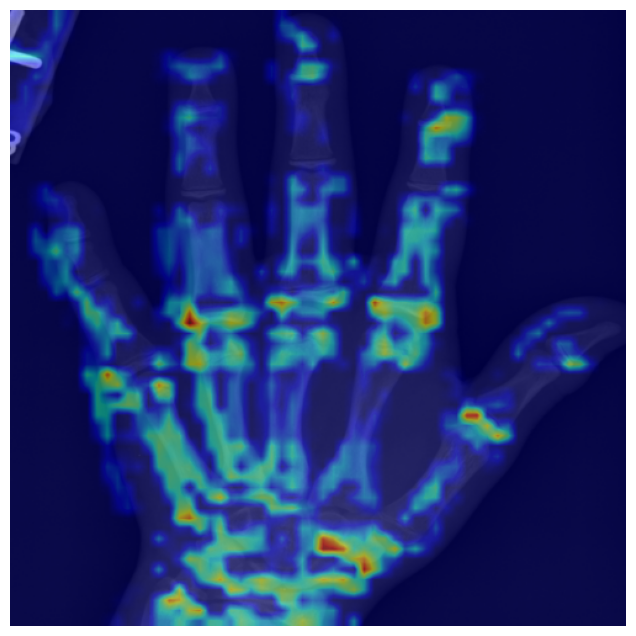} &
        \includegraphics[width=0.125\textwidth, height=2.2cm, trim=0 0 0 0, clip, keepaspectratio]{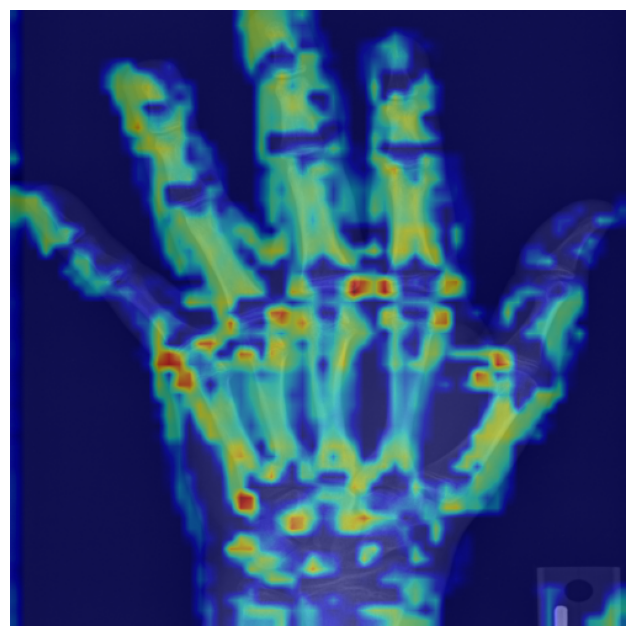} &
        \includegraphics[width=0.125\textwidth, height=2.2cm, trim=0 0 0 0, clip, keepaspectratio]{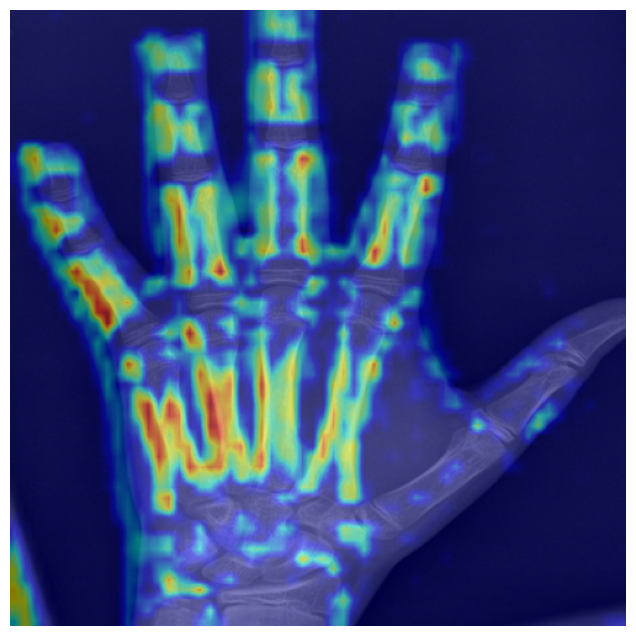} &
        \includegraphics[width=0.125\textwidth, height=2.2cm, trim=0 0 0 0, clip, keepaspectratio]{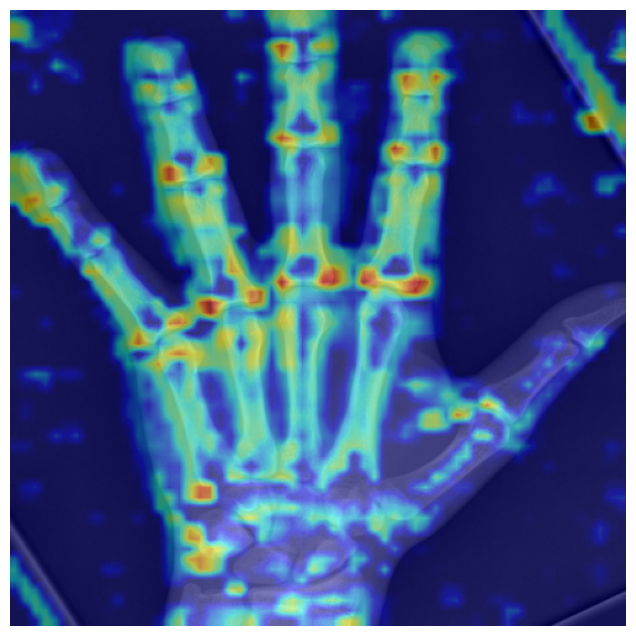} &
        \includegraphics[width=0.125\textwidth, height=2.2cm, trim=0 0 0 0, clip, keepaspectratio]{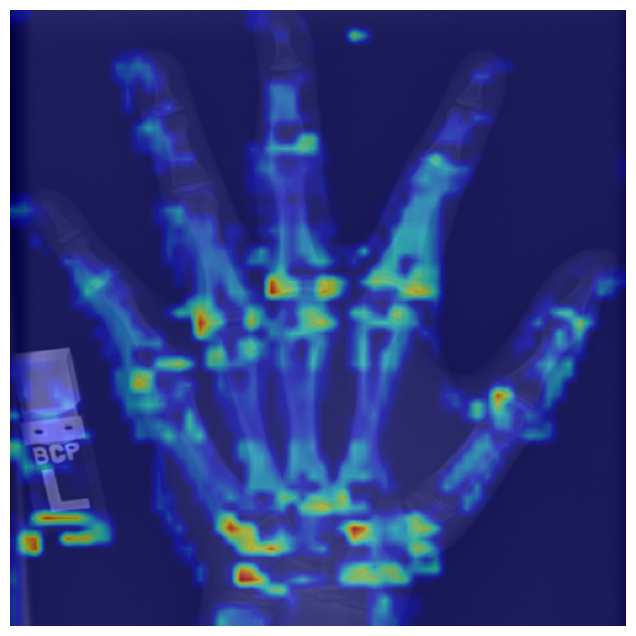} \\
        
        \includegraphics[width=0.125\textwidth, height=2.2cm, trim=0 0 0 0, clip, keepaspectratio]{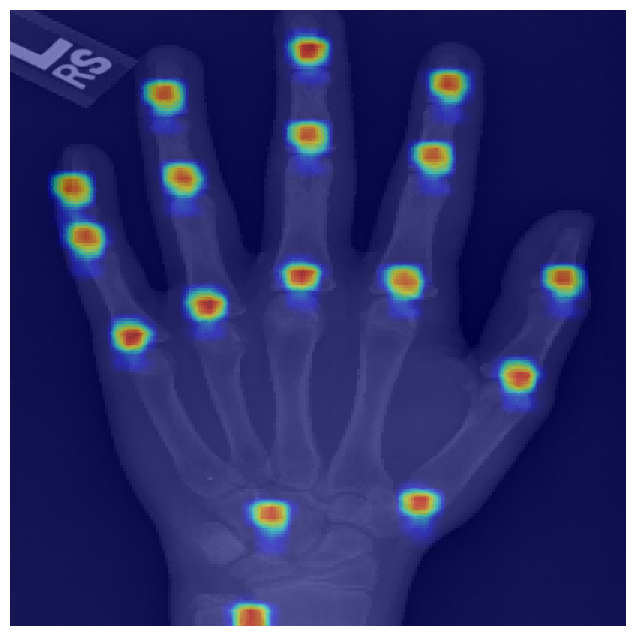} &
        \includegraphics[width=0.125\textwidth, height=2.2cm, trim=0 0 0 0, clip, keepaspectratio]{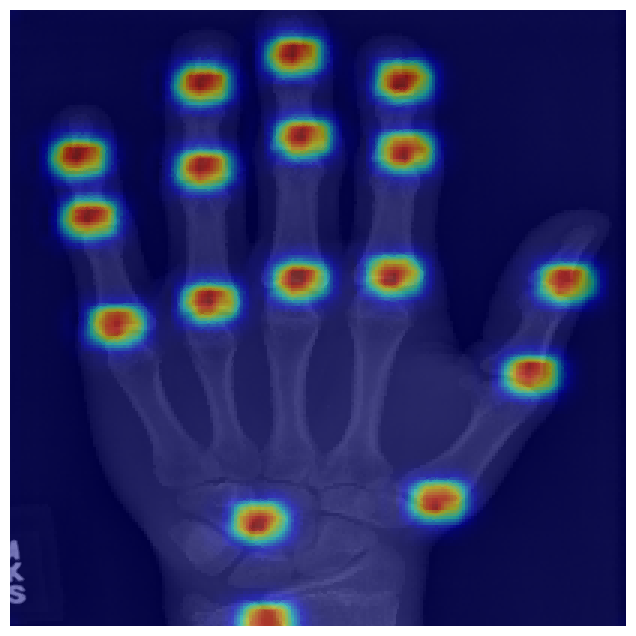} &
        \includegraphics[width=0.125\textwidth, height=2.2cm, trim=0 0 0 0, clip, keepaspectratio]{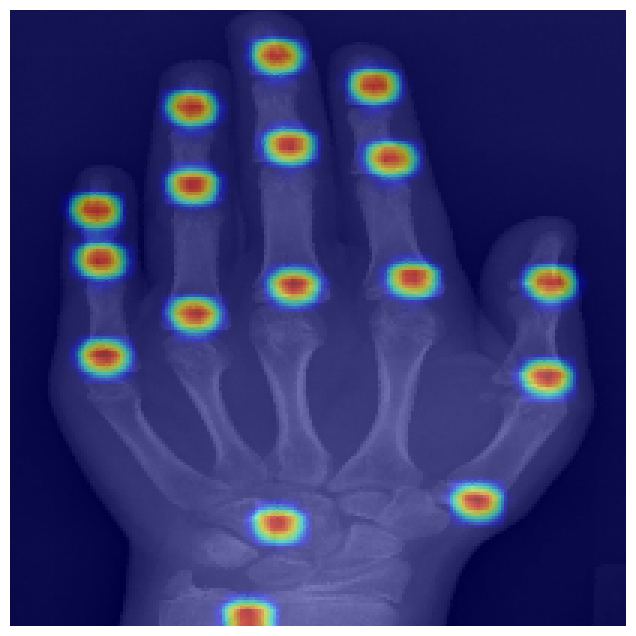} &
        \includegraphics[width=0.125\textwidth, height=2.2cm, trim=0 0 0 0, clip, keepaspectratio]{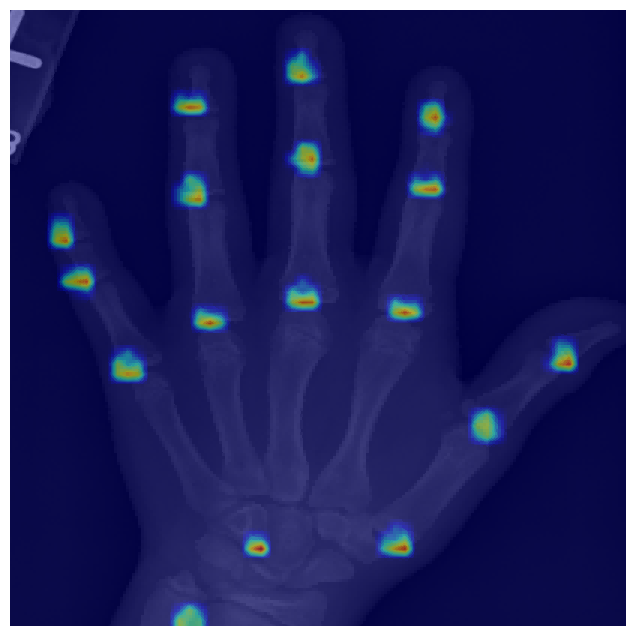} &
        \includegraphics[width=0.125\textwidth, height=2.2cm, trim=0 0 0 0, clip, keepaspectratio]{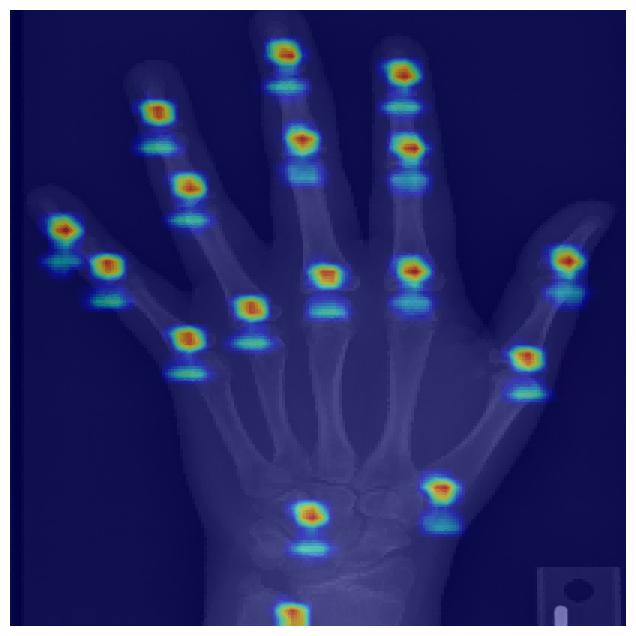} &
        \includegraphics[width=0.125\textwidth, height=2.2cm, trim=0 0 0 0, clip, keepaspectratio]{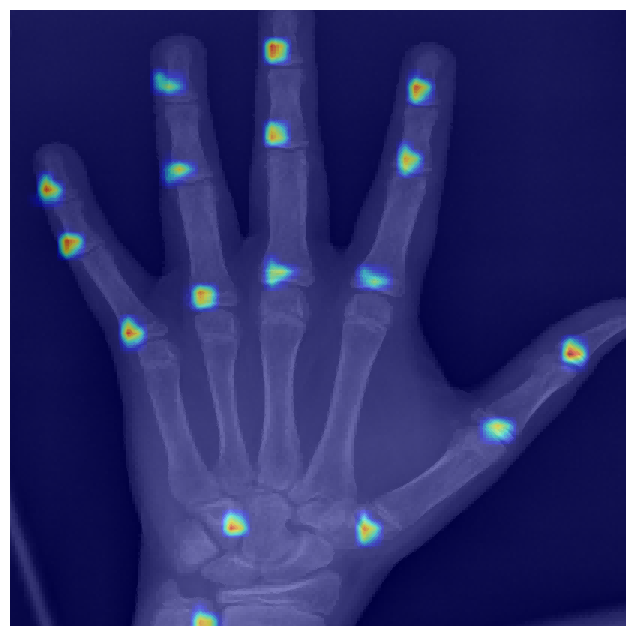} &
        \includegraphics[width=0.125\textwidth, height=2.2cm, trim=0 0 0 0, clip, keepaspectratio]{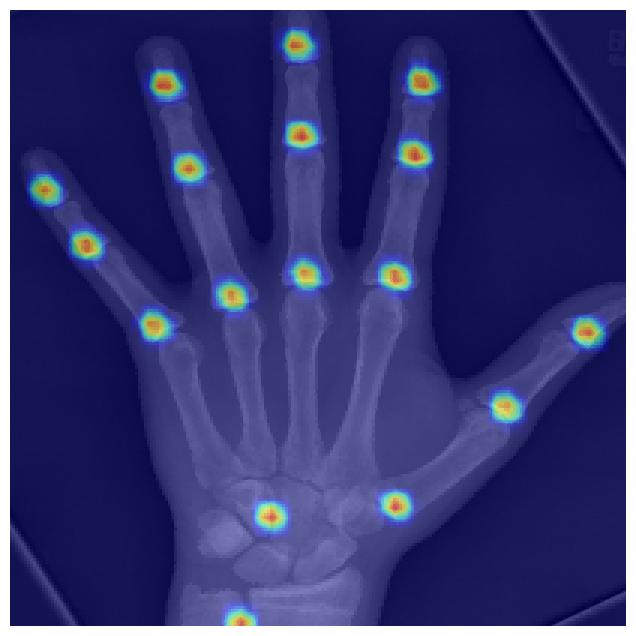} &
        \includegraphics[width=0.125\textwidth, height=2.2cm, trim=0 0 0 0, clip, keepaspectratio]{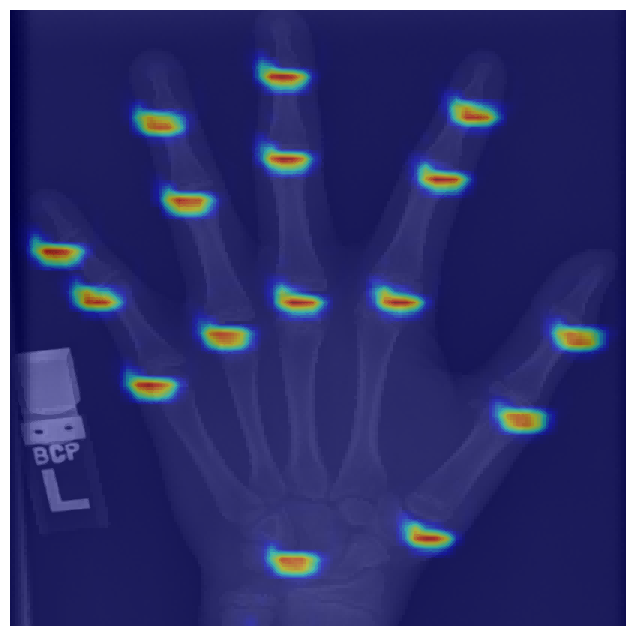} \\
        
        \includegraphics[width=0.125\textwidth, height=2.2cm, trim=0 0 0 0, clip, keepaspectratio]{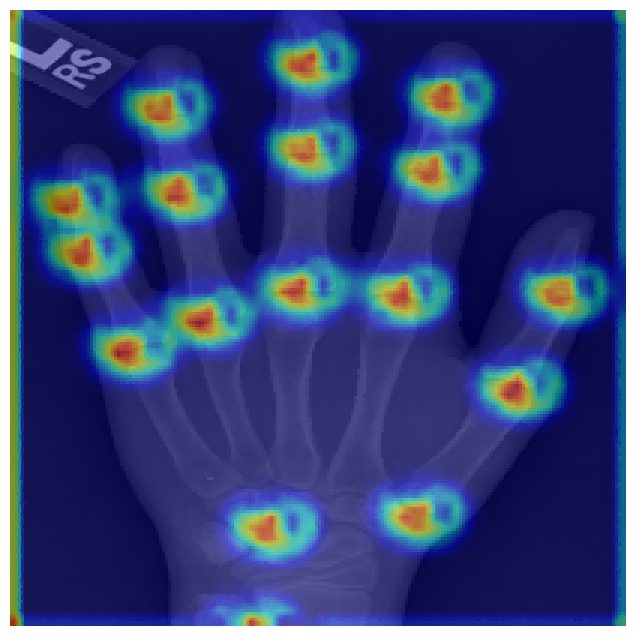} &
        \includegraphics[width=0.125\textwidth, height=2.2cm, trim=0 0 0 0, clip, keepaspectratio]{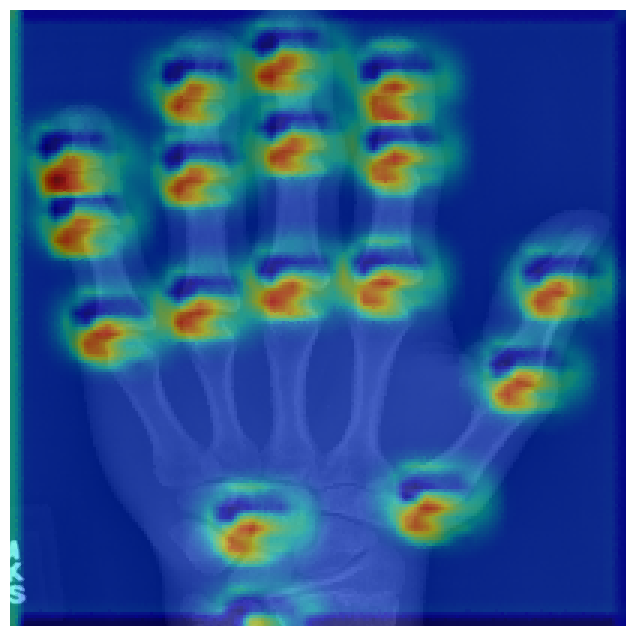} &
        \includegraphics[width=0.125\textwidth, height=2.2cm, trim=0 0 0 0, clip, keepaspectratio]{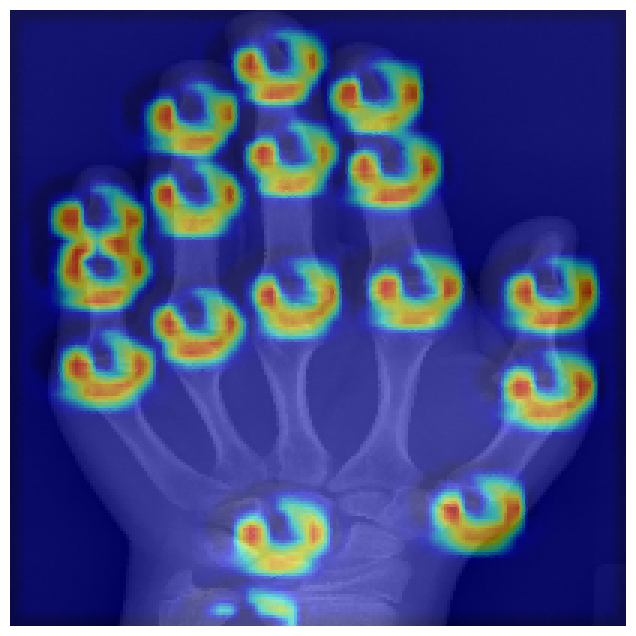} &
        \includegraphics[width=0.125\textwidth, height=2.2cm, trim=0 0 0 0, clip, keepaspectratio]{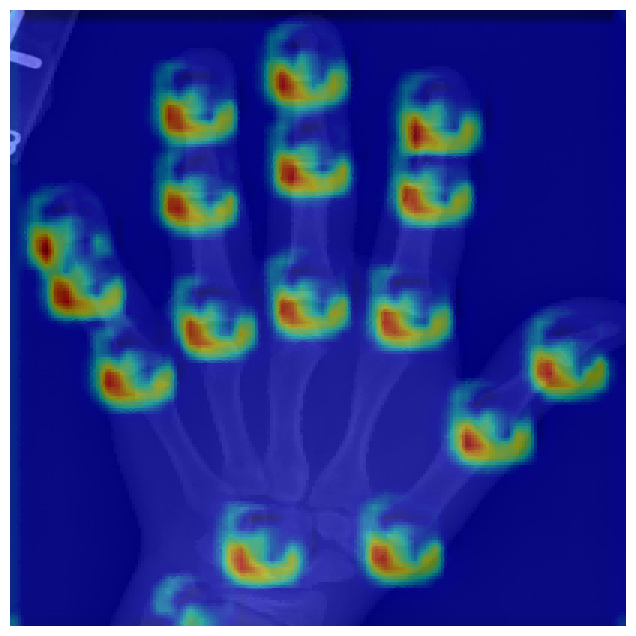} &
        \includegraphics[width=0.125\textwidth, height=2.2cm, trim=0 0 0 0, clip, keepaspectratio]{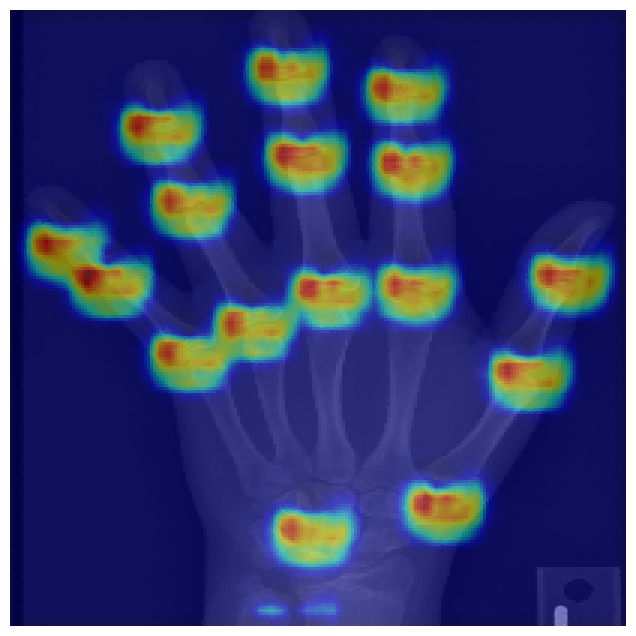} &
        \includegraphics[width=0.125\textwidth, height=2.2cm, trim=0 0 0 0, clip, keepaspectratio]{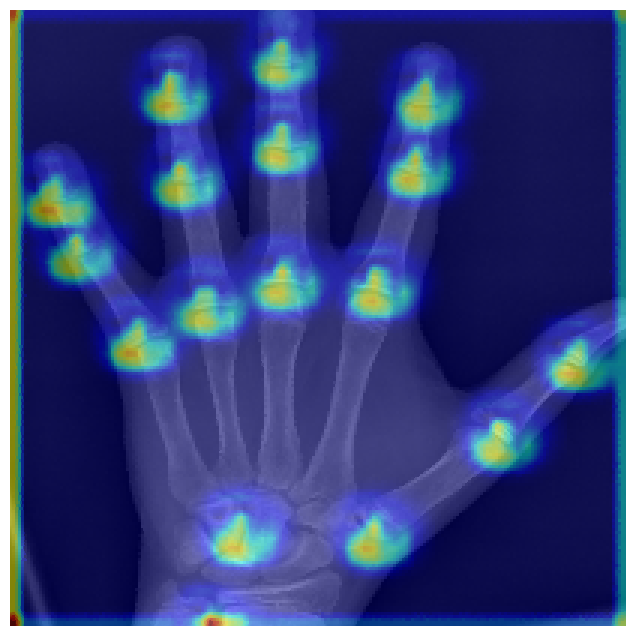} &
        \includegraphics[width=0.125\textwidth, height=2.2cm, trim=0 0 0 0, clip, keepaspectratio]{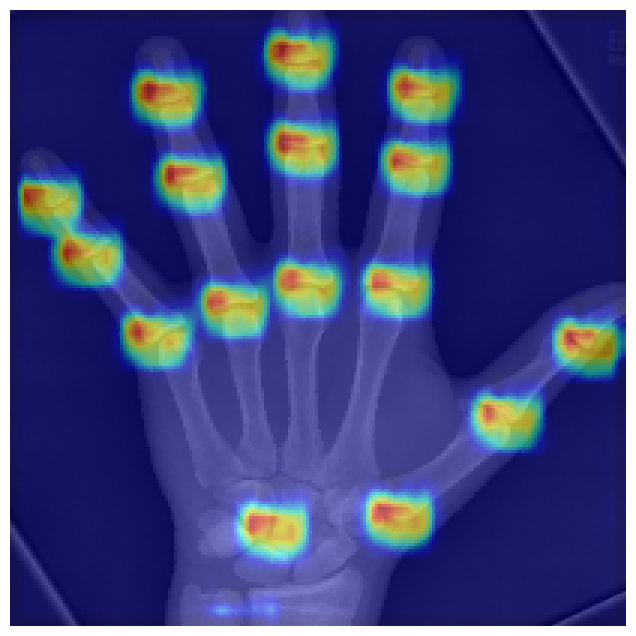} &
        \includegraphics[width=0.125\textwidth, height=2.2cm, trim=0 0 0 0, clip, keepaspectratio]{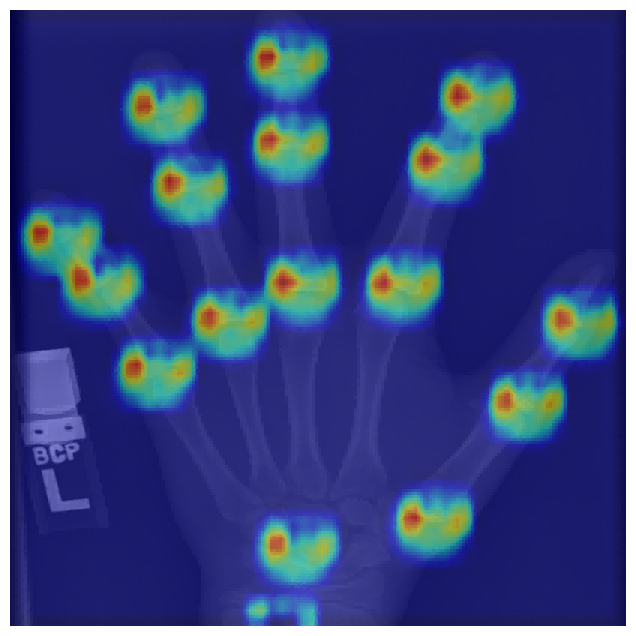} \\
    \end{tabular}
    \caption{Grad-CAM Visualization results on the RSNA dataset: The first row displays the whole hand radiographs. The second row displays the visualization of the feature map before processing by the Transformer module. The third row displays the visualization after processing by the Transformer module. The fourth row displays the visualization of the feature map before processing by the RFAConv module. The fifth row displays the visualization of the feature map after processing by the RFAConv module.}
    \label{fig:visualization}
\end{figure}

We employed Grad-CAM to visualize the function of the Transformer and RFAConv modules, comparing the interpretable heatmaps before and after each module to reveal the evolution of the model's focal regions. The results are shown in Fig.\ref{fig:visualization}. 

The second row shows the visualization of the feature map before the processing of the Transformer module. The model's focus is mainly on edge parts of the bones and the two ends of the bones, especially the terminal region of the proximal phalanges. However, the model has relatively weak grasp of the overall bone structure and difficulty in comprehensively capturing the key feature information of the small bone masses in the carpal bones. After being processed by the Transformer module (as shown in the third row), the model not only continues to focus on and strengthen the attention to the original key regions, but also extends its attention to the morphological outline of the entire hand bones, greatly enhancing the perception ability of the overall bone structure. Notably, the model can precisely focus on small bone masses (such as the trapezium, scaphoid, trapezoid, pisiform, and hamate bones) within the carpal, achieving accurate capture of these fine structures. These changes strongly demonstrate that the Transformer module has the ability to comprehensively and meticulously capture the details of the bone structure, laying a solid foundation for the model to more accurately assess the bone development stage subsequently. 

The fourth row shows the visualization of the feature map before the processing of the RFAConv module. The model's attention is concentrated around the 17 marked key points. In the carpal region, only one key point is set at the positions of the hamate and capitate bones, ignoring other important bones. In the phalangeal region, the key points are mainly marked at the ends of the distal, middle, and proximal phalanges. Consequently, the model only pays attention to individual bones, lacking focus on the connections between bones, which usually contain abundant development information. The metacarpal region is almost completely ignored by the model at this stage. After being processed by the RFAConv module (as shown in the fifth row), the model's attention region is greatly expanded. In the carpal region, the model not only strengthens its attention to the original key regions but also greatly expands its attention scope. The degree of attention to the hamate bone is greatly increased, the attention region to the capitate bone is greatly enlarged, and the previously neglected triquetral and lunate bones are successfully included in the attention scope. In the phalangeal and metacarpal regions, the model’s attention scope expands from the ends of individual bones to the epiphyseal regions in between phalanges or those connecting phalanges and metacarpal bones. This transformation enables the model to comprehensively analyze the associations and coordinated development relationships between different bones from an overall perspective, thereby more accurately and efficiently determining the bone development stage. The broader attention region makes the model more sensitive to changes in the size and shape of bones, enabling it to comprehensively and deeply capture the key information closely related to bone development. This fully demonstrates that the RFAConv module can effectively compensate for the limitations of insufficient key point marking, greatly enhancing the model's ability to capture information related to bone development. 

\section{Discussion}\label{sec:discussion}
While deep learning has advanced the field of bone age assessment in recent years, existing methods still face a core challenge: effectively balancing global skeletal features (overall developmental patterns) and local skeletal details (morphology of key bones). BoNet+ addresses this issue through a two-stream design, with its core logic closely aligning with the clinical reasoning of radiologists. Specifically, radiologists neither rely solely on the hand’s overall condition nor focus in isolation on individual key bones. Specifically, BoNet+ incorporates a Transformer module in the global feature extraction channel to capture overall skeletal development information from whole-hand radiographs, and integrates an RFAConv module in the local feature extraction channel to enhance local feature extraction capability and focus on key bones. Through this "global-local collaboration" mechanism, it achieves balanced utilization of information from both dimensions.  

Early black-box-based methods (e.g. Lee et al. \cite{ref7} and Larson et al. \cite{ref16}) process radiographs directly through a single network, often prioritizing global structures while neglecting local details. For instance, the CNN-based method by Lee et al. \cite{ref7} achieves MAE of 18.9 months on the RSNA test set, and the ResNet-based method by Larson et al. \cite{ref16} achieves MAE of 6.24 months on the same dataset. Both suffer from large errors due to their inability to fully capture subtle local skeletal features. In contrast, BoNet+'s two-stream structure fundamentally avoids this "trade-off" by explicitly separating and fusing global and local features. The ROI-based methods (e.g. Ji et al. \cite{ref23} and Mao et al. \cite{ref35}) perform well in extracting local details but tend to overlook global coordination between bone groups due to over-focus on predefined regions. Mao et al. \cite{ref35} segmented 18 ROIs using YOLOv5 and then extracted features via Swin Transformer. While this approach enhances local region analysis, it still results in an MAE of 4.586 months on the RSNA test set due to the lack of global context modeling. BoNet+ achieves a more balanced feature representation by fusing global skeletal relationships with local regional details, overcoming the limitations of such methods.  

Experimental results further validate the advantages of BoNet+: on the RSNA validation set, the cumulative accuracies within deviations of 6 months and 12 months are 72.8\% and 91.3\%, respectively; on the RHPE validation set, these metrics are 56.9\% and 82.6\%, respectively. These data not only indicate that BoNet+ yields smaller errors but also highlight its stability. BoNet+ reduces assessment variability through standardized global-local collaborative reasoning. This characteristic enables it to play an important role in high-pressure outpatient settings: providing physicians with rapid, reliable preliminary assessment references, reducing repetitive work, and allowing clinicians to focus more on accurate diagnosis of complex cases.

However, this study still has several limitations. First, the dataset exhibits an imbalance in the number of images across different age groups, which poses challenges for the generalization and application of the algorithm. Second, visualization results indicate that while the model focuses on the overall morphological structure of hand bones, it also shows some responsiveness to irrelevant background noise, such as lead markers and film edges, suggesting a limited ability to effectively filter out non-relevant information. This may introduce interference in complex clinical environments and compromise the model’s robustness.

Future research will focus on the following directions: first, expanding the dataset to balance image distribution across age groups and minimize evaluation bias; second, incorporating hand bone region segmentation to enhance assessment accuracy; third, exploring multimodal data fusion that integrates hand bone X-rays with ultrasound and magnetic resonance imaging, so as to provide new insights for BAA. These improvements are intended to better support clinical diagnosis, alleviate workload pressure in high-volume outpatient settings, and address the shortage of specialized physicians in underserved regions.

\section{Conclusions}\label{sec:conclusion}
Our paper introduces BoNet+, a two-stream deep learning framework for automated bone age assessment. This framework not only shortens diagnosis time and effectively reduces clinicians’ workload but also achieves high assessment accuracy. One stream of BoNet+ extracts global features, while the other extracts local features. In the global feature extraction channel, we introduce the Transformer module, which captures global skeletal development information from whole-hand radiographs. In the local feature extraction channel, we incorporate the RFAConv module, which effectively enhances local feature extraction capability and focuses on key bones. Then, by concatenating the features from these two channels and further processing the fused features through Inception-V3, the model fully leverages the advantages of combining both types of features. This integrated approach, which cooperatively combines the Transformer module's ability to capture global skeletal development information and the RFAConv module's ability to extract features of local key bones, enables BoNet+ to achieve good accuracy in bone age assessment. The framework breaks through the trade-off limitation of traditional single-channel models between global features and local details, and verifies the effectiveness of the global-local feature collaboration mechanism on multiple datasets. Experimental results show great improvements, with MAE reduced to 3.81 months on the RSNA test dataset and 5.65 months on the RHPE test dataset. Meanwhile, this approach reduces physicians' repetitive and mechanical tasks, optimizes the diagnostic and therapeutic workflow, provides reference support for clinical decision-making.

Although this study has limitations such as imbalanced age distribution of the dataset and insufficient noise resistance, the proposed two-stream model provides a promising solution for automated bone age assessment. Future research will further optimize the model by expanding and balancing the dataset, incorporating hand bone segmentation, and exploring multimodal data fusion, aiming to better support clinical diagnosis, alleviate the workload pressure in outpatient settings.
\vspace{6pt} 





\end{document}